%% file: main.tex
\documentclass[runningheads]{llncs}
\usepackage{graphicx}
\usepackage{comment}
\usepackage{amsmath,amssymb} 
\usepackage{color}

\usepackage{siunitx}
\usepackage{cite}
\usepackage{ctable}
\usepackage{hhline}
\usepackage{booktabs}
\usepackage{csquotes}
\usepackage{pifont}
\usepackage{multirow}
\usepackage{booktabs}
\usepackage{subcaption}
\usepackage{multicol}
\usepackage{hyperref}
\usepackage[super]{nth}

\newcolumntype{L}[1]{>{\raggedright\let\newline\\\arraybackslash\hspace{0pt}}m{#1}}
\newcolumntype{C}[1]{>{\centering\let\newline\\\arraybackslash\hspace{0pt}}m{#1}}
\newcolumntype{R}[1]{>{\raggedleft\let\newline\\\arraybackslash\hspace{0pt}}m{#1}}

\newcommand{\cmark}{\ding{51}}%
\newcommand{\xmark}{\ding{55}}%

\begin{document}
\pagestyle{headings}
\mainmatter

\title{InterHand2.6M: A Dataset and Baseline for \\ 3D Interacting Hand Pose Estimation \\ from a Single RGB Image}

\titlerunning{InterHand2.6M}
%
\author{Gyeongsik Moon \inst{1} \and Shoou-I Yu \inst{2} \and He Wen \inst{2} \and 
Takaaki Shiratori \inst{2} \and \\ Kyoung Mu Lee \inst{1}}
\authorrunning{G. Moon et al.}
%
\institute{ECE \& ASRI, Seoul National University, Korea \and
Facebook Reality Labs \\
\email{\{mks0601,kyoungmu\}@snu.ac.kr}, \email{\{shoou-i.yu,hewen,tshiratori\}@fb.com}}
\maketitle

\setcounter{footnote}{0} 

\begin{abstract}
Analysis of hand-hand interactions is a crucial step towards better understanding human behavior.
However, most researches in 3D hand pose estimation have focused on the isolated single hand case.
Therefore, we firstly propose (1) a large-scale dataset, InterHand2.6M, and (2) a baseline network, InterNet, for 3D interacting hand pose estimation from a single RGB image. The proposed InterHand2.6M consists of \textbf{2.6M labeled single and interacting hand frames} under various poses from multiple subjects. 
Our InterNet simultaneously performs 3D single and interacting hand pose estimation.
In our experiments, we demonstrate big gains in 3D interacting hand pose estimation accuracy when leveraging the interacting hand data in InterHand2.6M.
We also report the accuracy of InterNet on InterHand2.6M, which serves as a strong baseline for this new dataset.
Finally, we show 3D interacting hand pose estimation results from general images.
Our code and dataset are available\footnote{\url{https://mks0601.github.io/InterHand2.6M/}}.

\end{abstract}

\input{src/introduction.tex}

\input{src/related_works.tex}
\input{src/interhand2.6m_dataset.tex}

\input{src/internet.tex}
\input{src/implementation_details.tex}
\input{src/experiment.tex}

\input{src/conclusion.tex}

\noindent\textbf{Acknowledgments.}
We would like to thank 
Alexander Hypes, David Whitewolf, Eric Brockmeyer, Kevyn McPhail, Mark Pitts, Matt Stewart, Michal Perdoch, Scott Ardisson, Steven Krenn, and Timothy Godisart for building the capture system,
Autumn Trimble, Danielle Belko, Junko Saragih, Laura Millerschoen, Lucas Evans, Rohan Bali, Taylor Koska, and Xiaomin Luo for the data capture and annotation efforts, and
Chenglei Wu, Jason Saragih, Tomas Simon, and Yaser Sheikh for constructive feedback on data collection and the paper.
This work was partially supported by the Next-Generation Information Computing Development Program (NRF-2017M3C4A7069369) and the Visual Turing Test project (IITP-2017-0-01780) funded by the Ministry of Science and ICT of Korea.

\input{src/suppl.tex}

%
%
\bibliographystyle{splncs04}
\bibliography{main}
\end{document}

%% file: src/introduction.tex
\section{Introduction}

The goal of 3D hand pose estimation is to localize semantic keypoints (\textit{i.e.}, joints) of a human hand in 3D space. 
It is an essential technique for human behavior understanding and human-computer interaction. Recently, many methods~\cite{ge20173d,moon2018v2v,wan2019self,zimmermann2017learning,iqbal2018hand} utilize deep convolutional neural networks (CNNs) and have achieved noticeable performance improvement on public  datasets~\cite{tang2014latent,tompson2014real,sun2015cascaded,yuan20172017,zimmermann2017learning}.

\begin{figure}[t]
\begin{center}
\includegraphics[width=0.8\linewidth]{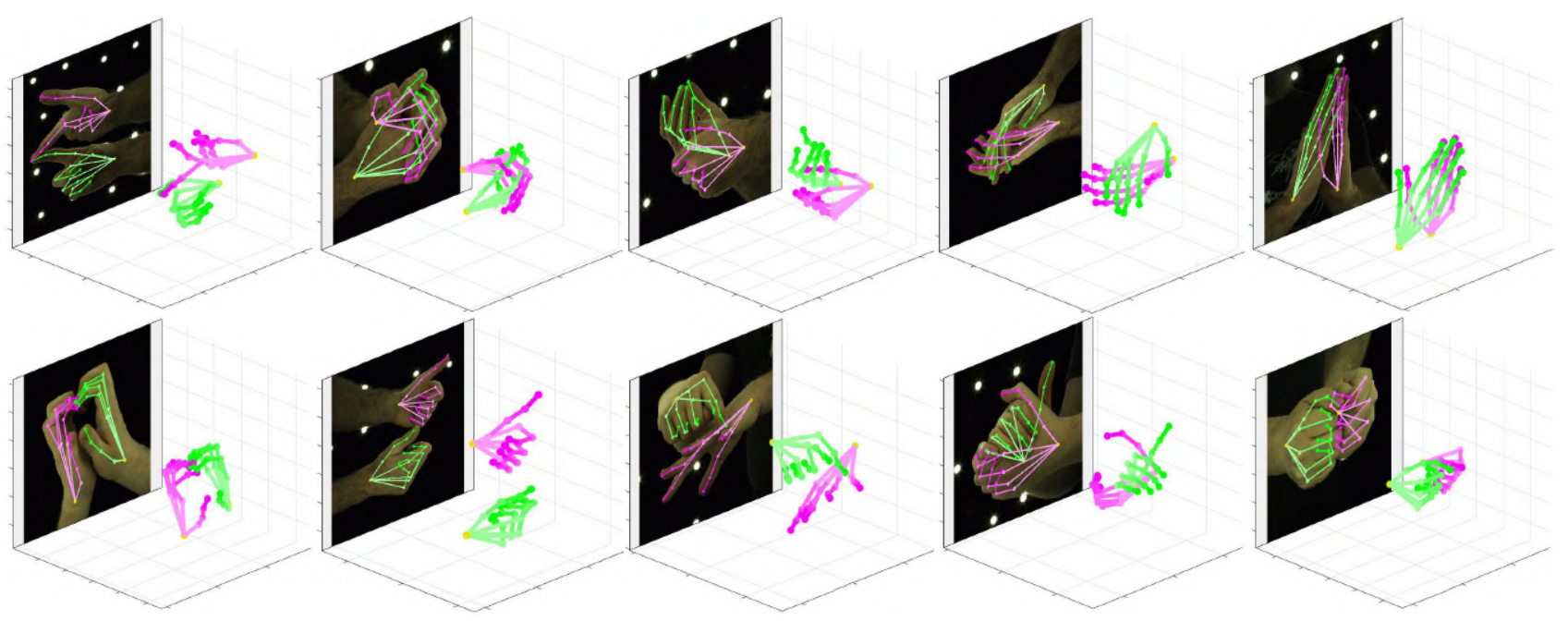}
\end{center}
   \caption{
Qualitative 3D interacting hand pose estimation results from our InterNet on the proposed InterHand2.6M.
   }
\label{fig:intro_qualitative}
\end{figure}

Most of the previous 3D hand pose estimation methods~\cite{ge20173d,moon2018v2v,wan2019self,zimmermann2017learning,iqbal2018hand} are designed for single hand cases. 
Given a cropped single hand image, models estimate the 3D locations of each hand keypoint. 
However, single hand scenarios have limitations in covering all realistic human hand postures because human hands often interact with each other to interact with other people and objects. 
To address this issue, we firstly propose a large-scale dataset, \emph{InterHand2.6M}, and a baseline, \emph{InterNet}, for 3D interacting hand pose estimation.

Our newly constructed InterHand2.6M is the first large-scale real (\textit{i.e.}, non-synthetic) RGB-based 3D hand pose dataset that includes both single and interacting hand sequences under various poses from multiple subjects.
Each hand sequence contains a single hand or interacting right and left hands of a single person. 
InterHand2.6M is captured in a precisely calibrated multi-view studio with 80 to 140 high-resolution cameras.
For 3D keypoint coordinate annotation, we use a semi-automatic approach, which is a combination of manual human annotation and automatic machine annotation. 
This approach makes annotation procedure much more efficient compared with full manual annotation while achieving similar annotation accuracy as the fully manual one.

The proposed InterNet simultaneously estimates 3D single and interacting hand pose from a single RGB image.
For this, we design InterNet to predict handedness, 2.5D right and left hand pose, and right hand-relative left hand depth.
The handedness can tell whether right or left hands are included in the input image; therefore InterNet can exclude the pose of a hand that does not exist in the testing stage.
The 2.5D hand pose consists of 2D pose in $x$- and $y$-axis and root joint (\textit{i.e.}, wrist)-relative depth in $z$-axis, widely used in state-of-the-art 3D human body~\cite{moon20193dmppe} and hand~\cite{iqbal2018hand} pose estimation from a single RGB image.
It provides high accuracy because of its image-aligned property and ability to model the uncertainty of the prediction.
To lift 2.5D right and left hand pose to 3D space, we obtain an absolute depth of the root joint from RootNet~\cite{moon20193dmppe}.
However, as obtaining absolute depth from a single RGB image is highly ambiguous, RootNet outputs unreliable depth in some cases.
To resolve this, we design InterNet to predict right hand-relative left hand depth by leveraging the appearance of the interacting hand from the input image.
This relative depth can be used instead of the output of the RootNet when both right and left hands are visible in the input image.

To demonstrate the benefit of the newly captured interacting hand data, we compare the performance of models trained on only single hand data, on only interacting hand data, and on both. 
We observed that models trained on interacting hand data achieve significantly lower interacting hand pose error than a model trained on single hand data.
This comparison shows that interacting hand data is essential for accurate 3D interacting hand pose estimation. 
We also demonstrate the effectiveness of our dataset for practical purposes by training InterNet on InterHand2.6M and showing its 3D interacting hand pose results from general images.
Figure~\ref{fig:intro_qualitative} shows 3D interacting hand pose estimation results from our InterNet on the proposed InterHand2.6M.

Our contributions can be summarized as follows.
\begin{itemize}
\item Our InterHand2.6M firstly contains large-scale high-resolution multi-view single and interacting hand sequences. 
By using a semi-automatic approach, we obtained accurate 3D keypoint coordinate annotations efficiently.

\item We propose InterNet for 3D single and interacting hand pose estimation. 
Our InterNet estimates handedness, 2.5D hand pose, and right hand-relative left hand depth from a single RGB image.

\item We show that single hand data is not enough, and interacting hand data is essential for accurate 3D interacting hand pose estimation.
\end{itemize}

%% file: src/related_works.tex
\section{Related works}

\noindent\textbf{Depth-based 3D single hand pose estimation.}
Early depth-based 3D hand pose estimation methods are mainly based on a generative approach. 
They fit a pre-defined hand model to the input depth image by minimizing hand-crafted cost functions~\cite{sharp2015accurate, tang2015opening}. Particle swarm optimization~\cite{sharp2015accurate}, iterative closest point~\cite{tagliasacchi2015robust}, and their combination~\cite{qian2014realtime} are the common algorithms used to obtain optimal hand poses.

Recent deep neural network-based methods are mainly based on a discriminative approach, which directly localizes hand joints from an input depth map.
Tompson~et al.~\cite{tompson2014real} firstly utilized the deep neural network to localize hand keypoints by estimating 2D heatmaps for each hand joint.
Ge~et al.~\cite{ge2016robust} extended this method by estimating multi-view 2D heatmaps.
Guo~et al.~\cite{guo2017ren} proposed a region ensemble network to estimate the 3D coordinates of hand keypoints accurately.
Moon~et al.~\cite{moon2018v2v} designed a 3D CNN model that takes voxel input and outputs a 3D heatmap for each keypoint.
Wan~et al.~\cite{wan2019self} proposed a self-supervised system, which can be trained only from an input depth map.

\begin{table}[t]
\centering
\setlength\tabcolsep{1.0pt}
\def\arraystretch{1.1}
\scalebox{1.0}{
\begin{tabular}{C{2.8cm}||C{2.0cm}C{1.9cm}C{1.7cm}C{0.8cm}C{0.9cm}C{1.3cm}}
\specialrule{.1em}{.05em}{.05em}
dataset & source  & resolution & annotation & sub. & fr. & int.hand \\ \hline
ICVL~\cite{tang2014latent} & real depth & 320$\times$240 & track & 10 & 18K & \xmark  \\
NYU~\cite{tompson2014real} & real depth & 640$\times$480 & track & 2 & 243K  & \xmark\\
MSRA~\cite{sun2015cascaded} & real depth & 320$\times$240 & track &  9 & 76K  & \xmark \\
BigHand2.2M~\cite{yuan2017bighand2} & real depth & 640$\times$480 & marker &  10 & 2.2M & \xmark \\
FPHA~\cite{garcia2018first}\footnote{There are markers on hands in the RGB sequence.} & real RGBD & 1920$\times$1080 & marker & 6 & 105K  & \xmark \\
Dexter+Object~\cite{sridhar2016real} & real RGBD & 640$\times$480 & manual & 1 & 3K  & \xmark \\
EgoDexter~\cite{mueller2017real} & real RGBD & 640$\times$480 & manual & 4 & 3K  & \xmark \\
STB~\cite{zhang20163d}  & real RGBD & 640$\times$480 & manual &  1 & 36K  & \xmark \\ 
FreiHAND~\cite{zimmermann2019freihand} & real RGB & 224$\times$224 & semi-auto. & 32 & 134K  & \xmark \\ \hline
RHP~\cite{zimmermann2017learning}  & synth. RGBD & 320$\times$320 & synth.  & 20 & 44K  & \xmark \\
Tzionas~et al.~\cite{tzionas2016capturing} & real RGBD & 640$\times$480 & manual & n/a & 36K  & \cmark \\
Mueller~et al.~\cite{mueller2019real} & synth. depth & n/a & synth. &  5 & 80K  & \cmark \\
Simon~et al.~\cite{simon2017hand} & real RGB & 1920$\times$1080 & semi-auto. &  n/a & 15K  & \cmark \\
\textbf{InterHand2.6M (ours)} & real RGB & \textbf{512$\times$334 (4096$\times$2668)} & semi-auto. & \textbf{27} & \textbf{2.6M}  & \cmark \\ \hline
\end{tabular}
}
\caption{Comparison of existing 3D hand pose estimation datasets and the proposed InterHand2.6M. 
For the RGBD-based datasets, we report their RGB resolution. 
For the multi-view captured datasets, we consider each image from different views as different images when reporting the number of frames. 
InterHand2.6M was initially captured at 4096$\times$2668 resolution, but to protect fingerprint privacy, the released set has resolution 512$\times$334.
}
\label{table:dataset_table_comparison}
\end{table}

\noindent\textbf{RGB-based 3D single hand pose estimation.}
Pioneering works~\cite{wu2005analyzing,de2011model} estimate hand pose from RGB image sequences.
Gorce~et al.~\cite{de2011model} proposed a model that estimates 3D hand pose, texture, and illuminant dynamically. 
Recently, deep learning-based methods show noticeable improvement. Zimmermann~et al.~\cite{zimmermann2017learning} proposed a deep neural network that learns a network-implicit 3D articulation prior.
Mueller~et al.~\cite{mueller2018ganerated} used an image-to-image translation model to generate a realistic hand pose dataset from a synthetic dataset.
Cai~et al.~\cite{cai2018weakly} and Iqbal~et al.~\cite{iqbal2018hand} implicitly reconstruct depth map and estimate 3D hand keypoint coordinates from it.
Spurr~et al.~\cite{spurr2018cross} and Yang~et al.~\cite{yang2019disentangling} proposed deep generative models to learn latent space for hand.

\begin{figure}[t]
\begin{center}
\includegraphics[width=0.7\linewidth]{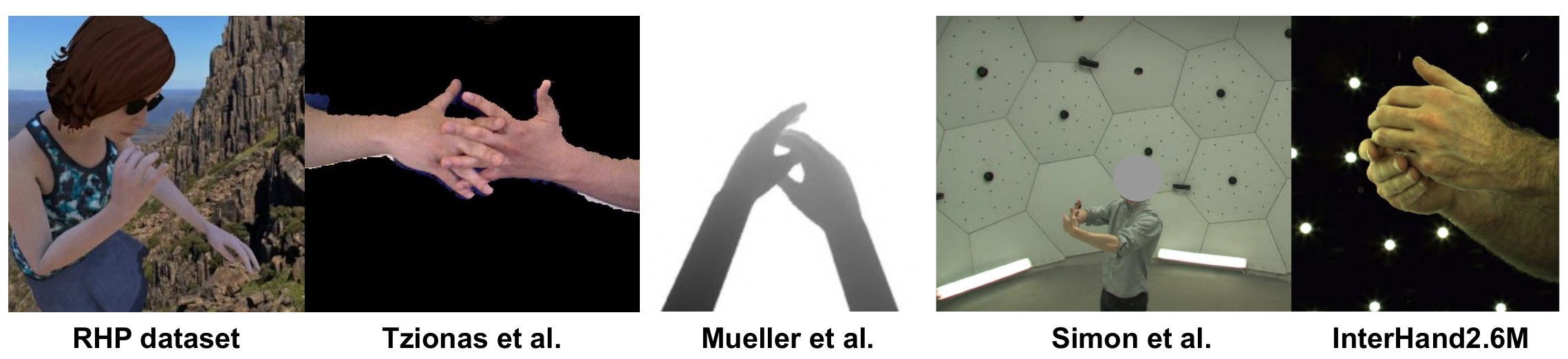}
\end{center}
   \caption{Comparisons of interacting hand images from RHP~\cite{zimmermann2017learning}, Tzionas~et al.~\cite{tzionas2016capturing}, Mueller~et al.~\cite{mueller2019real}, Simon~et al.~\cite{simon2017hand}, and the proposed InterHand2.6M.}
\label{fig:compare_db}
\end{figure}

\noindent\textbf{3D interacting hand pose estimation.}
There are few works that tried to solve the 3D interacting hand pose estimation. 
Oikonomidis~et al.~\cite{oikonomidis2012tracking} firstly attempted to address this problem using particle swarm optimization from an RGBD sequence.
Ballan~et al.~\cite{ballan2012motion} presented a framework that outputs 3D hand pose and mesh from multi-view RGB sequences.
They combined a generative model with discriminatively trained salient points to achieve a low tracking error.
Tzionas~et al.~\cite{tzionas2016capturing} extended Ballan~et al.~\cite{ballan2012motion} by incorporating a physical model.
Taylor~et al.~\cite{taylor2016efficient} proposed to perform joint optimization over both the hand model pose and the correspondences between observed data points and the hand model surface.
Simon~et al.~\cite{simon2017hand} performed 2D hand pose estimation from multi-view images and triangulated them into the 3D space.
Mueller~et al.~\cite{mueller2019real} proposed a model that estimates a correspondence map and hand segmentation map from a single depth map. 
The correspondence map provides a correlation between mesh vertices and image pixels, and the segmentation map separates right and left hand. 
They fit a hand model~\cite{romero2017embodied} to the estimated maps.

However, all of the above methods have limitations to be used for 3D single and interacting hand pose estimation from a single RGB image. 
Tzionas~et al.~\cite{tzionas2016capturing} and Simon~et al.~\cite{simon2017hand} require additional depth map or multi-view images.
The model of Mueller~et al.~\cite{mueller2019real} takes a single depth map and not a single RGB image.
In contrast, our proposed InterNet can perform 3D single and interacting hand pose estimation simultaneously from a single RGB image.

\noindent\textbf{3D hand pose estimation datasets.}
Table~\ref{table:dataset_table_comparison} shows specification of existing 3D hand pose datasets and the proposed InterHand2.6M. 
Compared with depth-based 3D hand pose estimation datasets~\cite{tang2014latent,tompson2014real,sun2015cascaded,yuan2017bighand2,garcia2018first}, existing RGB-based datasets~\cite{sridhar2016real,mueller2017real,zhang20163d,zimmermann2017learning} have very limited number of frames and subjects because obtaining accurate 3D annotation from RGB images is difficult.
Recently, Zimmermann~et al.~\cite{zimmermann2019freihand} captured a large-scale single hand pose and mesh dataset.

Several datasets contain two or interacting hand sequences, and Figure~\ref{fig:compare_db} shows example images of the datasets.
RHP~\cite{zimmermann2017learning} contains two isolated hand data.
However, their images are far from real because they are synthesized by animating 3D human models using commercial software. 
In addition, in most of their two hand images, right and left hands perform separate actions and are not interacting with each other.
The dataset of Tzionas~et al.~\cite{tzionas2016capturing} is the most similar dataset with ours in that it is constructed to analyze RGB interacting hand-focused sequences.
It contains RGBD interacting hand sequences, however only 2D joint coordinates annotations are available instead of the 3D coordinates.
In addition, the scale of the dataset is much smaller compared with that of our dataset.
The dataset of Mueller~et al.~\cite{mueller2019real} mainly consists of synthesized depth maps, which are not very realistic. 
Although some depth maps of their dataset are real-captured ones, 3D keypoint coordinate annotations of them are not available.
The dataset of Simon~et al.~\cite{simon2017hand} is not large-scale, and their annotations from a machine annotator are unstable because the resolution of the hand area of their dataset is low.

Compared with them, our InterHand2.6M consists of large-scale real-captured RGB images and includes more variety of sequences.
In addition, our strong machine annotator provides accurate and less jittering 3D hand joints coordinates annotations because of our strong semi-automatic annotation and high-resolution hand area.
Our dataset can be used when the hand is a central subject in the input image, for example, capturing the hand by a head-mounted device for virtual/augmented reality.

%% file: src/interhand2.6m_dataset.tex
\section{InterHand2.6M}

\subsection{Data Capture}

InterHand2.6M is captured in a multi-camera studio consisting of 80-140 cameras capturing at 30-90 frames-per-second (fps), and 350-450 directional LED point lights directed at the hand to promote uniform illumination\footnote{There were two settings. Setting 1: on average 34 RGB and 46 monochrome cameras (80 cameras total), 350 lights, and 90fps. Setting 2: on average 139 color cameras, 450 lights, and 30fps. Due to camera failures, not all cameras were operational; thus, each capture would have slightly different number of cameras.}.
The cameras captured at image resolution 4096 $\times$ 2668.
The multi-view system was calibrated with a 3D calibration target~\cite{ha2017deltille} and achieved pixel root mean square error ranging from 0.42 to 0.48.

We captured a total of 36 recordings consisting of 26 unique subjects, where 19 of them are males, and other 7 are females. 
There are two types of hand sequences\footnote{The examples of hand sequences are described in supplementary material.}.
First, peak pose (PP) is a short transition from neutral pose to pre-defined hand poses (\textit{e.g.}, fist) and then transition back to neutral pose. 
The pre-defined hand poses include various sign languages that are frequently used in daily life and extreme poses where each finger is maximally bent or extended.
There are 40 pre-defined hand poses for each right and left hand, and 13 for the interacting hand.
In the neutral pose, hands are in front of the person's chest, fingers do not touch, and palms face the side. 
The second type is a range of motion (ROM), which represents conversational gestures with minimal instructions. 
For example, subjects are instructed to wave their hands as if telling someone to come over. 
There are 15 conversational gestures for each right and left hand, and 17 for the interacting hand.
The hand poses from PP and ROM in our dataset are chosen to sample a variety of poses and conversational gestures while being easy to follow by capture participants.
The proposed InterHand2.6M is meant to cover a reasonable and general range of hand poses instead of choosing an optimal hand pose set for specific applications.

\subsection{Annotation}~\label{sec:annotation}
To annotate keypoints of hands, we directly extend the commonly used 21 keypoints per hand annotation scheme~\cite{zimmermann2017learning} to both hands, thus leading to a total of 42 unique points. 
For each finger, we annotate the fingertip and the rotation centers of three joints.
In addition to the 20 keypoints per hand, the wrist rotation center is also annotated.

Annotating rotation centers is challenging because the rotation center of a joint is occluded by the skin. 
The annotations become more challenging when the fingers are occluded by other fingers, or viewed from an oblique angle. 
Therefore, we developed a 3D rotation center annotation tool which allows the annotator to view and annotate 6 images simultaneously\footnote{The human annotation procedure is described in supplementary material.}. 
These 6 images are captured at the same time, but viewing the hand from different angles. 
When the annotator annotates a joint in two views, the tool will automatically perform triangulation and re-project the point to all other views, thus enabling the annotator to verify that the annotations are consistent in 3D space.

Despite having the annotation tool, manually annotating large amounts of images is still very labor-intensive.
Thus, we adopted a two-stage procedure to annotate the images following Simon~et al.~\cite{simon2017hand}. 
In the first stage, we rely on human annotators. 
The annotators leveraged our annotation tool and manually annotated 94,914 2D images from 9,036 unique time instants where 1,880 of them had two hand annotations.
These 2D annotations are triangulated to get 3D positions of joints, which are subsequently projected to all roughly 80 views to get 2D annotations for each view. 
The unique time steps are sampled to cover many hand poses of our recording scripts. 
At the end of this stage, a total of 698,922 images are labeled with 2D keypoints.

In the second stage, we utilize an automatic machine annotator. 
For this, we trained a state-of-the-art 2D keypoint detector~\cite{li2019rethinking} from the images annotated in the previous stage. EfficientNet~\cite{tan2019efficientnet} is used as a backbone of the keypoint detector for computational efficiency. 
The detector was then run through unlabeled images, and the 3D keypoints were obtained by triangulation with RANSAC. 
As our InterHand2.6M is captured from a large number of high-resolution cameras, this machine-based annotation gives highly accurate estimations. 
We tested this method on the held-out evaluation set, and the error is \textit{2.78 mm}. 
The final dataset is an integration of human annotations from the first stage and machine annotations from the second stage.
Simon~et al.~\cite{simon2017hand} performed iterative bootstrap because their initial machine annotator does not provide accurate annotations, and the hand area of their dataset has low resolution.
In contrast, our strong machine annotator on high-resolution hand images achieves significantly low error (2.78 mm); therefore, we did not perform iterative bootstrap.

\subsection{Dataset release}~\label{sec:annotation}

The captured hand sequences will be released under two configurations: downsized 512$\times$334 image resolution at 5 fps, and downsized 512$\times$334 resolution at 30 fps. Downsizing is to protect fingerprint privacy. 
The annotation file includes camera type, subject index, camera index, bounding box, handedness, camera parameters, and 3D joint coordinates.
All reported frame numbers and experimental results in the paper are from the 5 fps configuration.

%% file: src/internet.tex
\section{InterNet}

\begin{figure}[t]
\begin{center}
\includegraphics[width=0.7\linewidth]{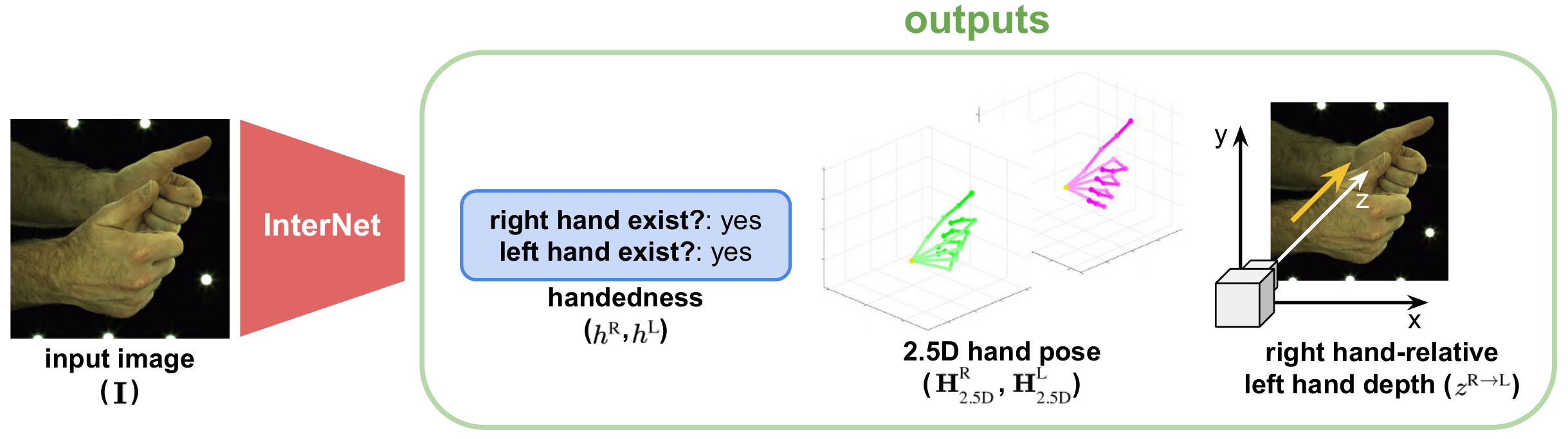}
\end{center}
   \caption{Three outputs of the proposed InterNet.}
\label{fig:overall_pipeline}
\end{figure}

Our InterNet takes a single RGB image $\mathbf{I}$ as an input and extracts the image feature $\mathbf{F}$ using ResNet~\cite{he2016deep} whose fully-connected layers are trimmed.
We prepare $\mathbf{I}$ by cropping the hand region from an image and resizing it to uniform resolution. 
From $\mathbf{F}$, InterNet simultaneously predicts handedness, 2.5D right and left hand pose, and right hand-relative left hand depth, which will be described in the following subsections.
We do not normalize the hand scale for the 2.5D hand pose estimation.
Figure~\ref{fig:overall_pipeline} shows overall pipeline of InterNet.

\subsection{Handedness estimation}
To decide which hand is included in the input image, we design our InterNet to estimate the probability of the existence of the right and left hand $\mathbf{h} = (h^\text R, h^\text L) \in \mathbb{R}^2$ in the input image.
For this, we build two fully-connected layers, which take the image feature $\mathbf{F}$ and estimates the probabilities $\mathbf{h}$. 
The hidden activation size of the fully-connected layers is 512.
Each fully-connected layer is followed by the ReLU activation function except for the last one. 
We apply a sigmoid activation function at the last layer to get the probabilities.

\subsection{2.5D right and left hand pose estimation}
To estimate 2.5D right and left hand pose, denoted as $\mathbf{P}^{\text R}_{\text {2.5D}} \in \mathbb{R}^{J \times 3}$ and $\mathbf{P}^{\text L}_{\text {2.5D}} \in \mathbb{R}^{J \times 3}$, respectively, we construct two upsamplers for each right and left hand. 
Each upsampler consists of three deconvolutional and one convolutional layers, and each deconvolutional layer is followed by batch normalization layers~\cite{ioffe2015batch} and ReLU activation functions, therefore it upsamples the input feature map $2^3$ times. 
The upsamplers take $\mathbf{F}$ and output 3D Gaussian heatmaps of the right and left hand joints, denoted as $\mathbf{H}^{\text R}_{\text{2.5D}}$ and $\mathbf{H}^{\text L}_{\text{2.5D}}$ following Moon~et al.~\cite{moon20193dmppe}, where they have the same dimension $\mathbb{R}^{J \times D \times H \times W}$.
$D$, $H$, and $W$ denote depth discretization size, height, and width of the heatmaps, respectively.
$x$- and $y$-axis of $\mathbf{H}^{\text R}_{\text{2.5D}}$ and $\mathbf{H}^{\text L}_{\text{2.5D}}$ are in image space, and $z$-axis of them are in root joint (\textit{i.e.}, wrist)-relative depth space. 
To obtain a 3D Gaussian heatmap from the 2D feature map, we reshape the output of the upsampler by a reshaping function $\psi \colon \mathbb{R}^{JD \times H \times W} \to \mathbb{R}^{J \times D \times H \times W}$.
Each voxel of the 3D Gaussian heatmap of the joint $j$ represents the likelihood of the existence of a hand joint $j$ in that position.

\subsection{Right hand-relative left hand depth estimation}
The depth of each hand is defined as that of the hand root joint. 
We construct two fully-connected layers and the ReLU activation function after each fully connected layer except for the last layer. 
The hidden activation size of the fully-connected layers is 512.
It takes $\mathbf{F}$ and outputs 1D heatmap $\mathbf{d}^{\text R \rightarrow \text L} \in \mathbb{R}^{64}$. 
Then, soft-argmax~\cite{sun2018integral} is applied to $\mathbf{d}^{\text R \rightarrow \text L}$ and output the relative depth value $z^{\text R \rightarrow \text L}$. 
We observed that estimating the 1D heatmap followed by soft-argmax operation provides a more accurate relative depth value compared with directly regressing it, which is a similar spirit to Moon~et al.~\cite{moon2018v2v}.

\subsection{Final 3D interacting hand pose}
The final 3D hand pose $\mathbf{P}^{\text R}_{\text{3D}}$ and $\mathbf{P}^{\text L}_{\text{3D}}$ are obtained as follows:
\begin{equation*}
\mathbf{P}^{\text R}_{\text{3D}}=\mathrm{\Pi}(\mathbf{T}^{-1}\mathbf{P}^{\text R}_{\text{2.5D}}+\mathbf{Z}^\text R),
\quad \text{and} \quad
\mathbf{P}^{\text L}_{\text{3D}}=\mathrm{\Pi}(\mathbf{T}^{-1}\mathbf{P}^{\text L}_{\text{2.5D}}+\mathbf{Z}^\text L),
\end{equation*}
where $\mathrm{\Pi}$ and $\mathbf{T}^{-1}$ denote camera back-projection and inverse affine transformation (\textit{i.e.}, 2D crop and resize), respectively.
We use normalized camera intrinsic parameters if not available following Moon~et al.~\cite{moon20193dmppe}.
$\mathbf{Z}^\text{R} \in \mathbb{R}^{1 \times 3}$ and $\mathbf{Z}^\text{L} \in \mathbb{R}^{1 \times 3}$ are defined as follows:

\noindent\begin{minipage}{.4\linewidth}
\begin{equation*}
\mathbf{Z}^\text R = [(0), (0), (z^\text R)],
\end{equation*}
\end{minipage}
\begin{minipage}{.5\linewidth}
\begin{equation*}
\mathbf{Z}^\text L= 
\begin{cases}
    [(0), (0), (z^\text L)],& \text{if } h^\text R < 0.5 \\
    [(0), (0), (z^\text{R}+z^{\text R \rightarrow \text L})],& \text{otherwise},
\end{cases}
~\label{eq:z_r_l}
\end{equation*}
\end{minipage}
\smallbreak
\noindent where $z^\text R$ and $z^\text L$ denote the absolute depth of the root joint of right and left hand, respectively.
We use RootNet~\cite{moon20193dmppe} to obtain them.

\subsection{Loss functions}
To train our InterNet, we use three loss functions.

\noindent\textbf{Handedness loss.}
For the handedness estimation, we use binary cross-entropy loss function as defined as follows:
\noindent$L_\text h = -\frac{1}{2} \sum_{\mathcal{Q} \in (\text R,\text L)} (\delta^\mathcal{Q} \log h^\mathcal{Q} + (1-\delta^\mathcal{Q})\log (1-h^\mathcal{Q}))$,
where $\delta^\mathcal{Q}$ is a binary value which represents existence of the $\mathcal{Q}$ hand in an input image.

\noindent\textbf{2.5D hand pose loss.}
For the 2.5D hand pose estimation, we use $L2$ loss as defined as follows:
\noindent$L_\text {pose} = \sum_{\mathcal{Q} \in (\text R,\text L)} ||\mathbf{H}^{\mathcal{Q}}_{\text{2.5D}} - \mathbf{H}^{\mathcal{Q}*}_{\text{2.5D}}||_2$,
where $*$ denotes groundtruth. 
If one of the right or left hand is not included in the input image, we set the loss from it zero.
The groundtruth 3D Gaussian heatmap is computed using a Gaussian blob~\cite{moon2018v2v} as follows:

\noindent$\mathbf{H}^{\mathcal{Q}*}_{\text{2.5D}}(j,z,x,y) =  \exp{\left(-\frac{(x-x_j^\mathcal{Q})^2+(y-y_j^\mathcal{Q})^2+(z-z_j^\mathcal{Q})^2}{2\sigma^2}\right)}$,
where $x_j^\mathcal{Q}$, $y_j^\mathcal{Q}$, and $z_j^\mathcal{Q}$ are $j$th joint coordinates of $\mathcal{Q}$ hand from $\mathbf{P}^\mathcal{Q}_{\text{2.5D}}$.

\noindent\textbf{Right hand-relative left hand depth loss.}
For the right hand-relative left hand localization, we use $L1$ loss as defined as follows:
\noindent$L_\text{rel} = |z^{\text R \rightarrow \text L} - z^{\text R \rightarrow \text L *}|$,
where $*$ denotes groundtruth.
The loss becomes zero when only a single hand is included in the input image.

We train our model in an end-to-end manner using all the three loss functions as follows:
$L = L_\text h + L_\text{pose} + L_\text{rel}$.

%% file: src/implementation_details.tex
\section{Implementation details}

PyTorch~\cite{paszke2017automatic} is used for implementation. 
The backbone part is initialized with the publicly released ResNet-50~\cite{he2016deep} pre-trained on the ImageNet dataset~\cite{russakovsky2015imagenet}, and the weights of the remaining part are initialized by Gaussian distribution with $\sigma=0.001$. 
The weights are updated by the Adam optimizer~\cite{kingma2014adam} with a mini-batch size of 64. 
To crop the hand region from the input image, we use groundtruth bounding box in both of training and testing stages. 
The cropped hand image is resized to 256$\times$256; thus the spatial size of the heatmap is $H \times W = 64 \times 64$.
We set $D=64$.
Data augmentations including translation ($\pm$15\%), scaling ($\pm$25\%), rotation ($\pm$\ang{90}), horizontal flip, and color jittering ($\pm$20\%) is performed in training.
The initial learning rate is set to $10^{-4}$ and reduced by a factor of 10 at the \nth{15} and \nth{17} epoch. 
We train our model for 20 epochs with four NVIDIA TitanV GPUs, which takes 48 hours when training on our InterHand2.6M. 
Our InterNet runs at a speed of 53 fps.

%% file: src/experiment.tex
\section{Experiment}

\subsection{Dataset and evaluation metric}

\noindent\textbf{STB.} 
STB~\cite{zhang20163d} includes 6 pairs of stereo sequences of diverse poses with different backgrounds from a single person. 
For evaluation, end point error (EPE) is widely used, which is defined as a mean Euclidean distance (mm) between the predicted and ground-truth 3D hand pose after root joint alignment.

\noindent\textbf{RHP.} 
RHP~\cite{zimmermann2017learning} has a large number of synthesized images.
They used 3D human models of 20 different subjects to synthesize 39 actions. For the evaluation metric, EPE is used.

\noindent\textbf{InterHand2.6M.} InterHand2.6M is our newly captured 3D hand pose dataset. 
We split our InterHand2.6M into training, validation, and test set, as shown in Table~\ref{table:training_validation_test_set}. 
Val (M) and Test (M) contain many unseen hand poses and only subjects not seen in Train (H+M). 
Also, Val (M) and Test (M) only consists of ROM, which includes longer and more diverse sequences than that of Train (H+M).
This can make Val (M) and Test (M) more similar to real-world scenarios. 
Test (H) contains many seen hand poses, and half of the subjects are seen in Train (H). 
There are duplicated frames and annotations in Train (H) and Train (M), and we overwrite them with Train (H).

\begin{table}[t]
\centering
\setlength\tabcolsep{1.0pt}
\def\arraystretch{1.1}
\scalebox{1.0}{
\begin{tabular}{C{2.5cm}|C{1.7cm}C{1.5cm}C{2.0cm}C{2.0cm}C{2.0cm}}
\specialrule{.1em}{.05em}{.05em}
split & sequence & subjects & frames (SH) & frames (IH) & frames (All) \\ \hline
Train (H) & PP+ROM & 16 & 142K  & 386K &  528K \\
Train (M) & PP+ROM & 9 & 594K  & 315K & 909K  \\
Train (H+M) & PP+ROM & 21 & 688K  & 674K & \textbf{1361K}  \\ \hline
Val (M) & ROM & 1 & 234K & 146K & \textbf{380K} \\ \hline
Test (H) & PP+ROM & 6 & 34K & 88K & 122K \\ 
Test (M) & ROM & 2 & 455K & 272K & 728K \\
Test (H+M) & PP+ROM & 8 & 489K & 360K & \textbf{849K} \\ \hline
\end{tabular}
}
\caption{Training, validation, and test set split of the proposed InterHand2.6M. H and M denote human annotation and machine annotation, respectively. SH and IH denote single and interacting hand, respectively.}
\label{table:training_validation_test_set}
\end{table}

For the evaluation, we introduce three metrics. 
First, we use the average precision of handedness estimation ($\mathrm{AP_h}$) to measure the accuracy of handedness estimation. 
Second, the mean per joint position error (MPJPE) is defined as a Euclidean distance (mm) between predicted and groundtruth 3D joint locations after root joint alignment.
The root joint alignment is performed for each left and right hand separately.
This metric measures how accurately the root-relative 3D hand pose estimation is performed. 
Last, mean relative-root position error (MRRPE) is defined as a Euclidean distance (mm) between predicted and groundtruth right hand root-relative left hand root position. 
It measures how right hand-relative left hand localization is accurately performed. 

\subsection{Ablation study}

\noindent\textbf{Benefit of the interacting hand data.}
To investigate the benefit of the interacting hand data for 3D interacting hand pose estimation, we compare single and interacting hand MPJPE of our InterNet trained with and without interacting hand data in Table~\ref{table:sh_ih_training}.
For all settings, we used the same RootNet trained on both single and interacting hand data. 
As the table shows, a model trained only on interacting hand provides significantly lower interacting hand pose error than a model trained only on single hand data. 
This shows existing 3D single hand pose estimation datasets are not enough for accurate 3D interacting hand pose estimation. 
In addition, we trained a model on combined single and interacting hand data, which is our InterHand2.6M. 
We observed that additional interacting hand data improves not only interacting hand pose estimation performance, but also single hand pose estimation. 
These comparisons clearly show the benefit of our newly introduced interacting hand data for 3D single and interacting hand pose estimation.

\begin{table}[t]
\setlength{\tabcolsep}{1pt}
\begin{minipage}{.5\linewidth}
\centering

\scalebox{1.0}{
\begin{tabular}{C{2.4cm}C{1.7cm}C{1.7cm}}
\specialrule{.1em}{.05em}{.05em}
training set & SH MPJPE & IH MPJPE  \\ \hline
SH only & 13.08 & 51.19    \\
IH only & 13.70 & 16.86   \\
\textbf{SH+IH (ours)}  & \textbf{12.16} & \textbf{16.02}  \\ \hline
\specialrule{.1em}{.05em}{.05em}
\end{tabular}
}
\caption{Single and interacting hand MPJPE comparison from models trained on the different training sets. 
SH and IH denote single and interacting hand, respectively.}
\label{table:sh_ih_training}
\end{minipage}\hfill
\begin{minipage}{.50\linewidth}
\centering

\begin{center}
   \includegraphics[width=0.7\linewidth]{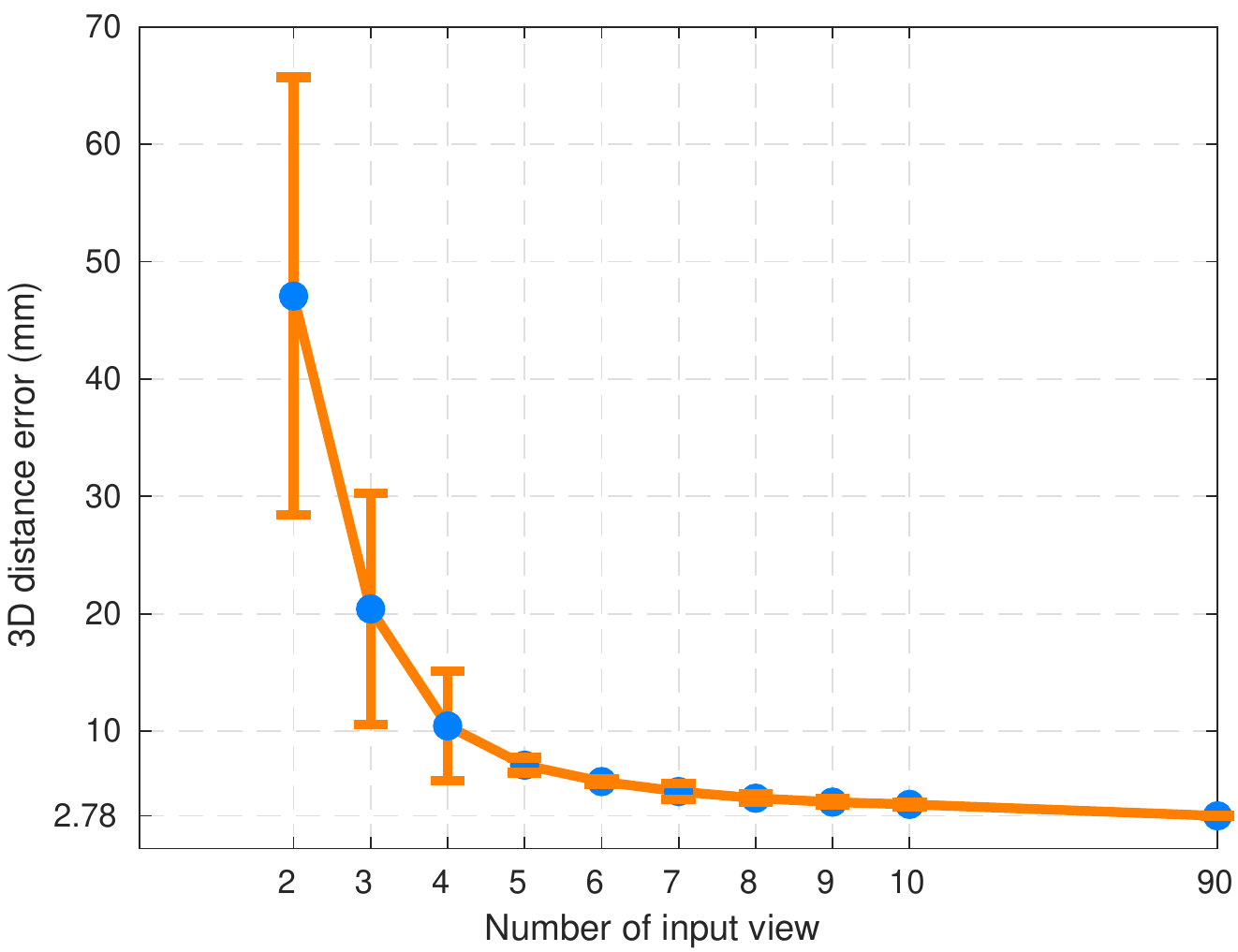}
\end{center}
   \captionof{figure}{The 3D distance error of our machine annotation on the Test (H). 
   }
\label{fig:accuracy_of_machine}
\end{minipage}
\end{table}

\begin{table}[t]
\centering
\setlength\tabcolsep{1.0pt}
\def\arraystretch{1.1}
\scalebox{1.0}{
\begin{tabular}{C{2.5cm}C{2.0cm}C{2.0cm}C{2.0cm}C{2.0cm}}
\specialrule{.1em}{.05em}{.05em}
training set & Val (M) & Test (H) & Test (M) & Test (H+M) \\ \hline
Train (H) & 15.02/19.70 & 10.42/13.05 & 12.74/18.10 & 12.58/17.16\\
Train (M) & 15.36/20.13 & 10.64/14.26 & 12.56/18.59  & 12.43/17.79 \\
\textbf{Train (H+M)}  & \textbf{14.65}/\textbf{18.58} & \textbf{9.85}/\textbf{12.29} & \textbf{12.32}/\textbf{16.88} & \textbf{12.16}/\textbf{16.02}\\ \hline
\end{tabular}
}
\caption{Single and interacting hand MPJPE comparison from models trained on the different training sets. 
The numbers on the left of the slash are single hand, and the ones on the right are interacting hand MPJPE.}
\label{table:benefit_of_machine_annot}
\end{table}

\noindent\textbf{Accuracy of the machine annotation.}
To show the accuracy of our machine annotation model, we train our annotation model on Train (H) and test on Test (H).
Figure~\ref{fig:accuracy_of_machine} shows 3D distance error (mm) on Test (H) according to the number of input views. 
For each number of input views, the vertical line represents a standard deviation, which shows performance variation due to view selection.
In the testing time, the model takes randomly selected $v$ views and performs 2D hand pose estimation, followed by triangulation. 
To cover various combinations of selecting $v$ views from all $V$ views, we repeat the same testing procedure 100 times for each $v$ views. 
The figure shows that as the number of input views increases, both the error and standard deviation becomes smaller, and finally, the error becomes \textit{2.78 mm} when all 90 views are used. 
This shows our annotation method is highly accurate by utilizing state-of-the-art 2D keypoint detection network and a large number of views for triangulation.

\noindent\textbf{Benefit of machine-generated annotation.}
To show the benefit of the automatically obtained machine annotations, we compare the accuracy of models trained without and with Train (M) in Table~\ref{table:benefit_of_machine_annot}. 
As the table shows, a model trained on Train (H) achieves better performance than a model trained on Train (M). 
We hypothesize that although our machine annotator has very low 3D distance error, human annotation is still more accurate, which makes a model trained on Train (H) performs better. 
However, as the machine provides annotation more efficiently than a human, it can annotate many frames easily that may not be included in Train (H). 
Therefore, this machine-generated annotations can have better coverage of hand pose space, which can be a benefit in the training stage. 
To utilize this better coverage, we add machine annotation to the human annotation. 
The last row of the table shows that a model trained on the combined dataset achieves the best accuracy. 
This comparison clearly shows the benefit of adding the machine-generated annotation to the human annotation. 
We provide more analysis of human and machine annotation in the supplementary material.

\noindent\textbf{Benefit of using $z^{\text R \rightarrow \text L}$.}
To show the benefit of using $z^{\text R \rightarrow \text L}$ when right hand is visible (\textit{i.e.}, $h^\text R \ge 0.5$) instead of always using $z^\text L$, we compare $\mathrm{MRRPE}$ between the two cases.
We checked that $\mathrm{MRRPE}$ of always using $z^\text L$ is 92.14 mm, while that of using $z^{\text R \rightarrow \text L}$ when $h^\text R \ge 0.5$ is 32.57 mm.
This is because estimating $z^\text L$ from a cropped single image inherently involves high depth ambiguity because the camera position is not provided in the cropped input image.
In contrast, estimating $z^{\text R \rightarrow \text L}$ from a cropped image involves less depth ambiguity because both hands are visible in the cropped input image.

\begin{table}[t]
\centering
\setlength\tabcolsep{1.0pt}
\def\arraystretch{1.1}
\scalebox{1.0}{
\begin{tabular}{C{3.5cm}C{0.8cm}C{0.9cm}C{1.6cm}C{1.7cm}}
\specialrule{.1em}{.05em}{.05em}
methods & GT S & GT H & EPE (STB) & EPE (RHP) \\ \hline
Zimm.~et al.~\cite{zimmermann2017learning} & \cmark & \cmark & 8.68 & 30.42\\
Chen~et al.~\cite{chen2018generating} & \cmark & \cmark & 10.95 & 24.20\\
Yang~et al.~\cite{yang2019disentangling}  & \cmark & \cmark & 8.66 & 19.95  \\
Spurr~et al.~\cite{spurr2018cross} & \cmark & \cmark & 8.56 & 19.73  \\ 
Spurr~et al.~\cite{spurr2018cross} & \xmark & \xmark & 9.49 & 22.53 \\ 
\textbf{InterNet (ours)} & \xmark & \xmark & \textbf{7.95} & \textbf{20.89} \\ \hline
\end{tabular}
}
\caption{EPE comparison with previous state-of-the-art methods on STB and RHP. 
The checkmark denotes a method use groundtruth information during inference time. 
S and H denote scale and handness, respectively.}
\label{table:stb_rhp_comparison_with_sota}
\end{table}

\subsection{Comparison with state-of-the-art methods}
We compare the performance of our InterNet with previous state-of-the-art 3D hand pose estimation methods on the STB and RHP in Table~\ref{table:stb_rhp_comparison_with_sota}.
The table shows the proposed InterNet outperforms previous methods without relying on ground-truth information during inference time. 
Our InterNet estimates 3D heatmap of each joint, while other methods directly estimate 3D joint coordinates. 
As shown in Moon~et al.~\cite{moon2018v2v}, directly regressing 3D joint coordinates from an input image is a highly non-linear mapping. 
In contrast, our InterNet estimates per-voxel likelihoods, which makes learning easier and provides state-of-the-art performance.

\subsection{Evaluation on InterHand2.6M} 
Table~\ref{table:benefit_of_machine_annot} shows 3D errors of InterNet on InterHand2.6M. 
Table~\ref{table:sh_ih_training} shows that InterNet trained on both single and interacting hand data yields the 32\% larger error on interacting hand sequences than single hand sequences. 
This comparison tells us that interacting hand sequences are harder to analyze than single hand cases.
To analyze the difficulty of InterHand2.6M, we compare our error with 3D hand pose error of current state-of-the-art depth map-based 3D hand pose estimation methods~\cite{moon2018v2v,xiong2019a2j} on the large-scale depth map 3D hand pose datasets~\cite{yuan2017bighand2,Yuan_2018_CVPR}. 
They achieved $8\sim 9$ mm error on large scale depth map dataset~\cite{yuan2017bighand2,Yuan_2018_CVPR}, which is far less than 3D interacting hand pose estimation error of our InterNet (\textit{i.e.}, 16.02 mm). 
Considering our InterNet achieves state-of-the-art performance on publicly available datasets~\cite{zhang20163d,zimmermann2017learning}, we can conclude that 3D interacting hand pose estimation from a single RGB image is far from solved.
Our InterNet achieves 99.09 $\mathrm{AP_h}$ and 32.57 $\mathrm{MRRPE}$ on Test (H+M).

\begin{figure}[t]
\begin{center}
\includegraphics[width=0.7\linewidth]{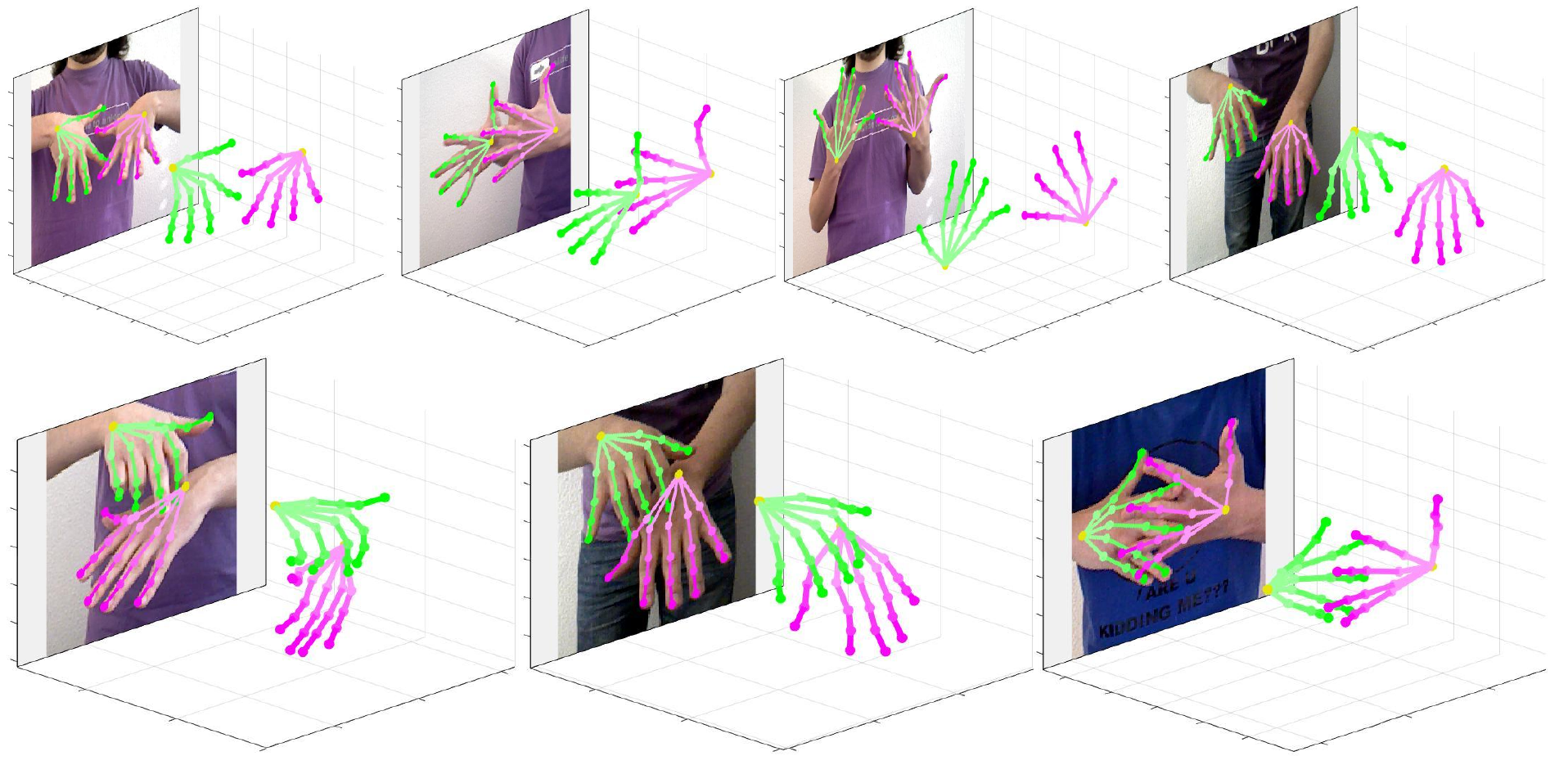}
\end{center}
   \caption{Qualitative results on the dataset of Tzionas~et al.~\cite{tzionas2016capturing}, which is captured from a general environment.}
\label{fig:general_images}
\end{figure}

\subsection{3D interacting hand pose estimation from general images}
We show 3D interacting hand pose estimation results from general images in Figure~\ref{fig:general_images}.
For this, we additionally utilize the dataset of Tzionas~et al.~\cite{tzionas2016capturing}, which is captured from the general environment but only provides the 2D groundtruth joints coordinates.
We randomly split the dataset of Tzionas~et al.~\cite{tzionas2016capturing} at a 9:1 ratio as a training and testing set, respectively.
During the training, a mini-batch consists of half-InterHand2.6M and half-dataset of Tzionas~et al.~\cite{tzionas2016capturing}.
For the simultaneous 3D and 2D supervision from our dataset and that of Tzionas~et al.~\cite{tzionas2016capturing}, respectively, we perform soft-argmax~\cite{sun2018integral} on the estimated heatmaps $\mathbf{H}^\text R$ and $\mathbf{H}^\text L$ to extract the 3D coordinates in a differentiable way.
Then, we modify $L_\text{pose}$ to a $L1$ distance between the extracted 3D coordinates and the groundtruth.
We set a loss of $z$-axis coordinate to zero when the input image is from the dataset of Tzionas~et al.~\cite{tzionas2016capturing}.
The figure shows our InterNet successfully produces 3D interacting hand pose results from general images from the dataset of Tzionas~et al.~\cite{tzionas2016capturing}, although the 3D supervision is only applied to the data from our InterHand2.6M.

%% file: src/conclusion.tex
\section{Conclusion}

We propose a baseline, InterNet, and dataset, InterHand2.6M, for 3D interacting hand pose estimation from a single RGB image. 
The proposed InterHand2.6M is the first large-scale 3D hand pose dataset that includes various single and interacting hand sequences from multiple subjects.
As InterHand2.6M only provides 3D hand joint coordinates, fitting 3D hand model~\cite{romero2017embodied} to our dataset for the 3D rotation and mesh data of interacting hand can be interesting future work.

%% file: src/suppl.tex
\clearpage

\begin{center}
\textbf{\large Supplementary Material of \enquote{InterHand2.6M: \\ A Dataset and Baseline for \\ 3D Interacting Hand Pose Estimation \\ from a Single RGB Image}}
\end{center}

In this supplementary material, we present more experimental results that could not be included in the main manuscript due to the lack of space.

\section{Comparison of human and machine annotation}
\begin{figure}
\begin{center}
\includegraphics[width=0.5\linewidth]{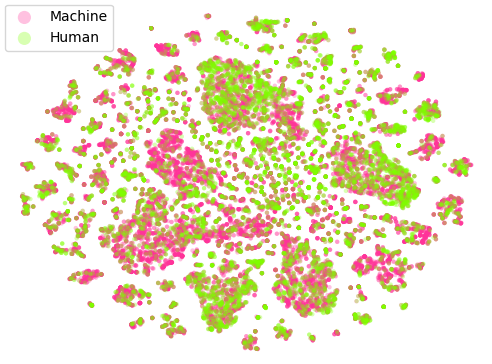}
\end{center}
\vspace*{-5mm}
   \caption{Visualized t-SNE of human and machine-generated 3D hand pose annotation.}
\vspace*{-3mm}
\label{fig:tsne_machine_human}
\end{figure}

To show how human and machine annotations are different, we visualize t-SNE of human and machine annotation in Figure~\ref{fig:tsne_machine_human}. Each vector in t-SNE is a 20-dimensional hand pose vector. For this, we pre-defined 20 degrees of freedom for each hand. Two degrees of freedom are defined for each finger root (\textit{i.e.}, T1, I1, M1, R1, and P1) as pitch and yaw angles. The other degree of freedom is defined for T2, T3, I2, I3, M2, M3, R2, R3, P2, and P3 as pitch angle. To calculate the angles, we assume I1, M1, R1, P1, and wrist joints are on the same plane $s$. As the figure shows, the machine-generated annotations have broader hand pose coverage than human-generated ones.

\section{Effect of number of the available views}
Our InterHand2.6M dataset has a large number of views. To show how the available number of views affect the 3D hand pose estimation accuracy, we report MPJPE of a model trained only from four widely used views (\textit{i.e.}, top, frontal, right, and left views) in Table~\ref{table:input_view_result}. For the fair comparison, we increased the number of iterations when using four views for training to make the total number of iterations in the training stage the same. Also, the same RootNet~\cite{moon20193dmppe} trained on all views is used for both settings. As the table shows, our large number of views improves the performance significantly. This comparison shows a large number of views in our InterHand2.6M dataset is beneficial.

\begin{table}
\centering
\setlength\tabcolsep{1.0pt}
\def\arraystretch{1.1}
\scalebox{0.8}{
\begin{tabular}{C{3.4cm}||C{1.5cm}C{1.5cm}C{1.5cm}}
\specialrule{.1em}{.05em}{.05em}
num. of training views & $\mathrm{AP_{h}}$ & MRRPE & MPJPE \\ \hline
4 & 0.95 & 61.90 & 26.80 \\
\textbf{all (ours)} & \textbf{0.99} & \textbf{32.57} & \textbf{14.22} \\ \hline
\end{tabular}
}
\caption{$\mathrm{AP_{h}}$, MRRPE, and MPJPE on Test (H+M) of InterHand2.6M dataset using different number of training views.}
\vspace*{-3mm}
\label{table:input_view_result}
\end{table}


\section{InterHand2.6M sequence descriptions}
We provide detailed descriptions and visualizations of the sequences in the proposed InterHand2.6M dataset.

\begin{figure}
\begin{center}
\includegraphics[width=1.0\linewidth]{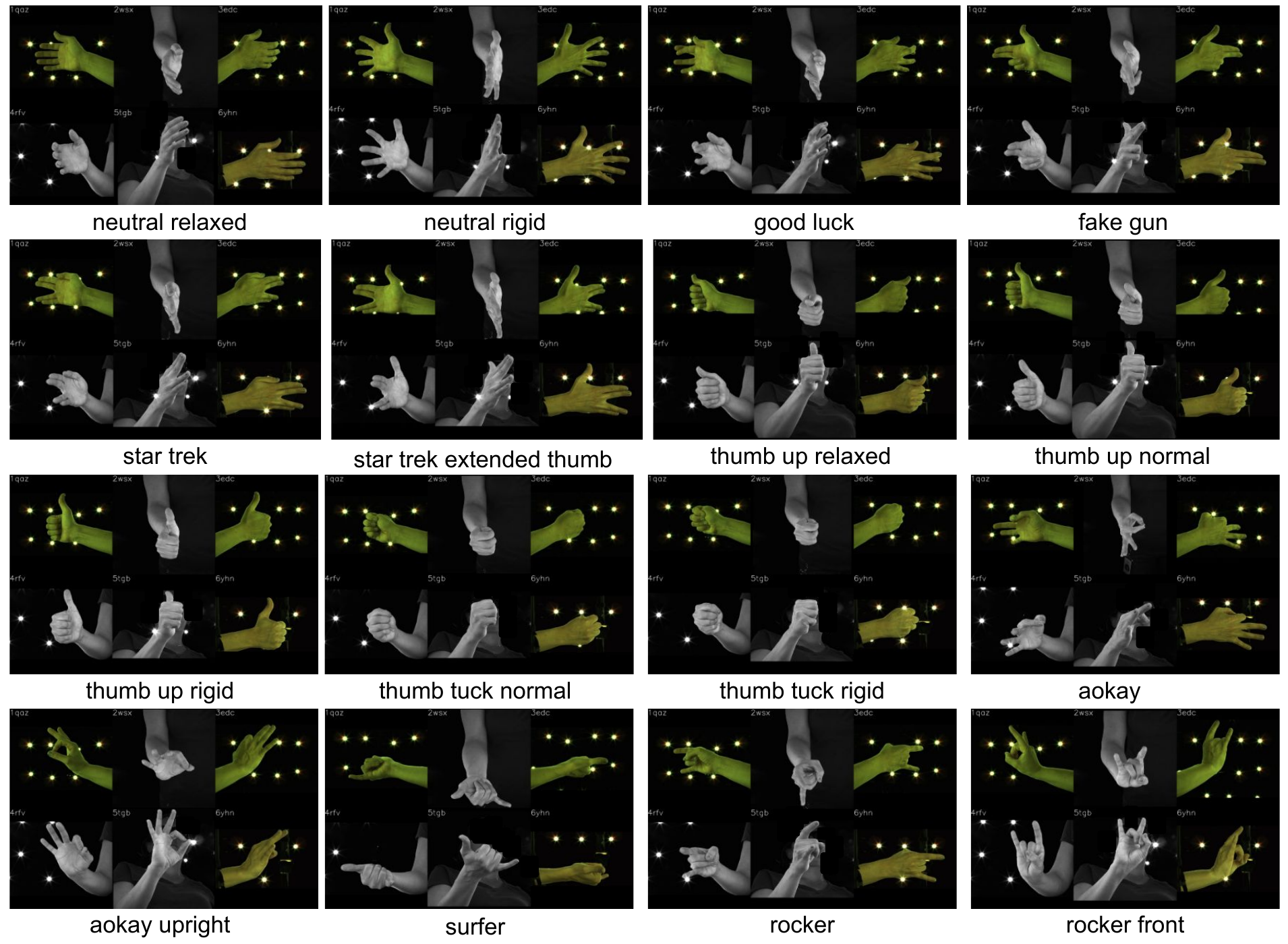}
\end{center}
\vspace*{-5mm}
   \caption{Visualization of the single hand PP sequences.}
\vspace*{-3mm}
\label{fig:single_pp_1}
\end{figure}

\begin{figure}
\begin{center}
\includegraphics[width=1.0\linewidth]{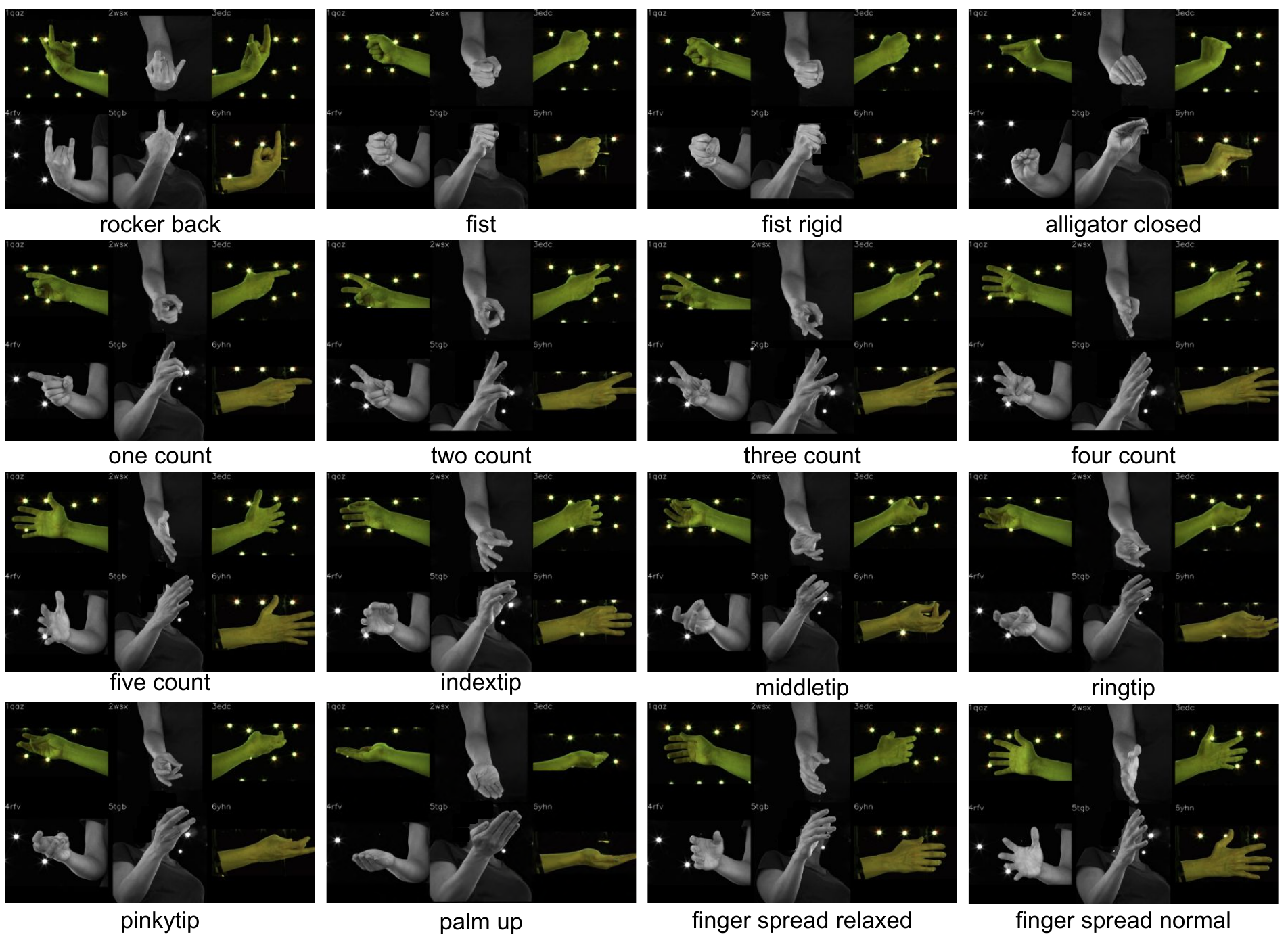}
\end{center}
\vspace*{-5mm}
   \caption{Visualization of the single hand PP sequences.}
\vspace*{-3mm}
\label{fig:single_pp_2}
\end{figure}

\begin{figure}
\begin{center}
\includegraphics[width=1.0\linewidth]{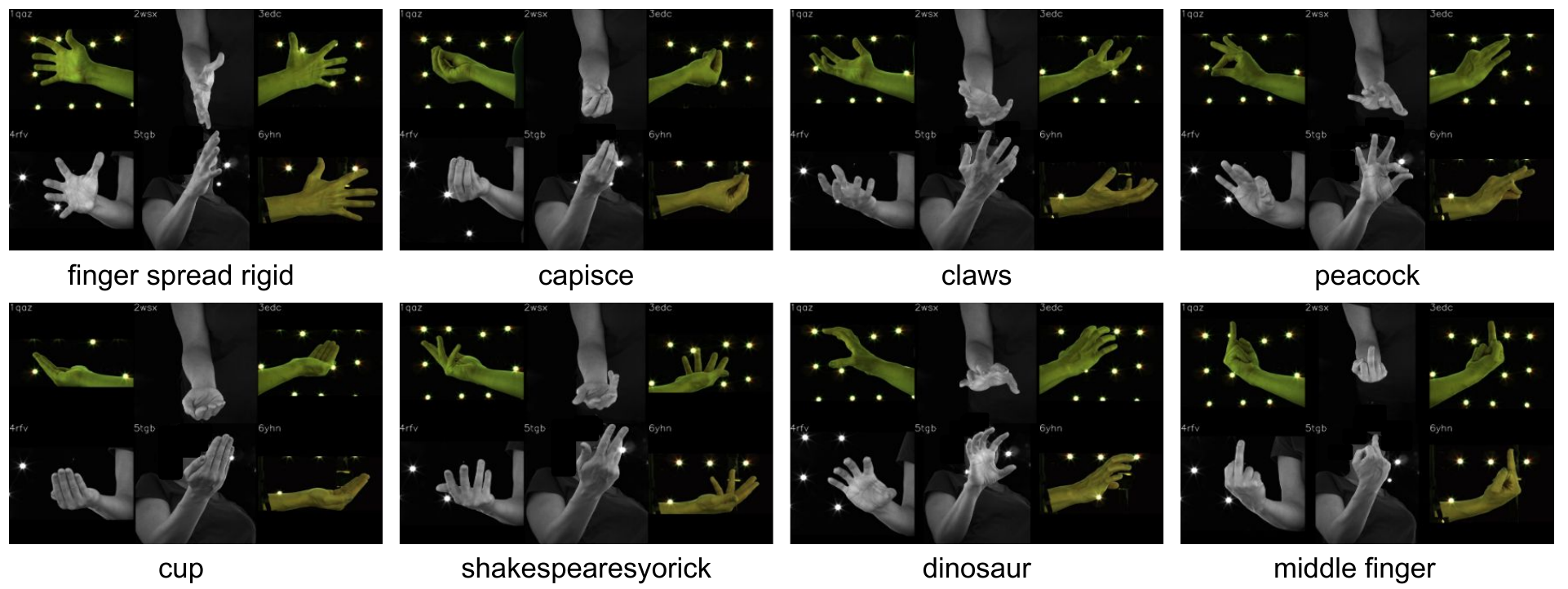}
\end{center}
\vspace*{-5mm}
   \caption{Visualization of the single hand PP sequences.}
\vspace*{-3mm}
\label{fig:single_pp_3}
\end{figure}

\noindent\textbf{Single hand sequences.} 
Figure~\ref{fig:single_pp_1}, ~\ref{fig:single_pp_2}, and ~\ref{fig:single_pp_3} show PP of single hand sequences. 
Below are detailed descriptions of each sequence.
\begin{itemize}
\item neutral relaxed: the neutral hand pose. Hands in front of the chest, fingers do not touch, and palms face the side.
\item neutral rigid: the neutral hand pose with maximally extended fingers, muscles tense.
\item good luck: hand sign language with crossed index and middle fingers.
\item fake gun: hand gesture mimicking the gun.
\item star trek: hand gesture popularized by the television series Star Trek.
\item star trek extended thumb: \enquote{star trek} with extended thumb.
\item thumb up relaxed: hand sign language that means \enquote{good}, hand muscles relaxed.
\item thumb up normal: \enquote{thumb up}, hand muscles average tenseness.
\item thumb up rigid: \enquote{thumb up}, hand muscles very tense.
\item thumb tuck normal: similar to fist, but the thumb is hidden by other fingers.
\item thumb tuck rigid: \enquote{thumb tuck}, hand muscles very tense.
\item aokay: hand sign language that means \enquote{okay}, where palm faces the side.
\item aokay upright: \enquote{aokay} where palm faces the front.
\item surfer: the SHAKA sign.
\item rocker: hand gesture that represents rock and roll, where palm faces the side.
\item rocker front: the \enquote{rocker} where palm faces the front.
\item rocker back: the \enquote{rocker} where palm faces the back.
\item fist: fist hand pose.
\item fist rigid: fist with very tense hand muscles.
\item alligator closed: hand gesture mimicking the alligator with a closed mouth.
\item one count: hand sign language that represents \enquote{one.}
\item two count: hand sign language that represents \enquote{two.}
\item three count: hand sign language that represents \enquote{three.}
\item four count: hand sign language that represents \enquote{four.}
\item five count: hand sign language that represents \enquote{five.}
\item indextip: thumb and index fingertip are touching.
\item middletip: thumb and middle fingertip are touching.
\item ringtip: thumb and ring fingertip are touching.
\item pinkytip: thumb and pinky fingertip are touching.
\item palm up: has palm facing up.
\item finger spread relaxed: spread all fingers, hand muscles relaxed.
\item finger spread normal: spread all fingers, hand muscles average tenseness.
\item finger spread rigid: spread all fingers, hand muscles very tense.
\item capisce: hand sign language that represents \enquote{got it} in Italian.
\item claws: hand pose mimicking claws of animals.
\item peacock: hand pose mimicking peacock.
\item cup: hand pose mimicking a cup.
\item shakespeareyorick: hand pose from Yorick from Shakespeare's play Hamlet.
\item dinosaur: hand pose mimicking a dinosaur.
\item middle finger: hand sign language that has an offensive meaning.
\end{itemize}

\begin{figure}
\begin{center}
\includegraphics[width=1.0\linewidth]{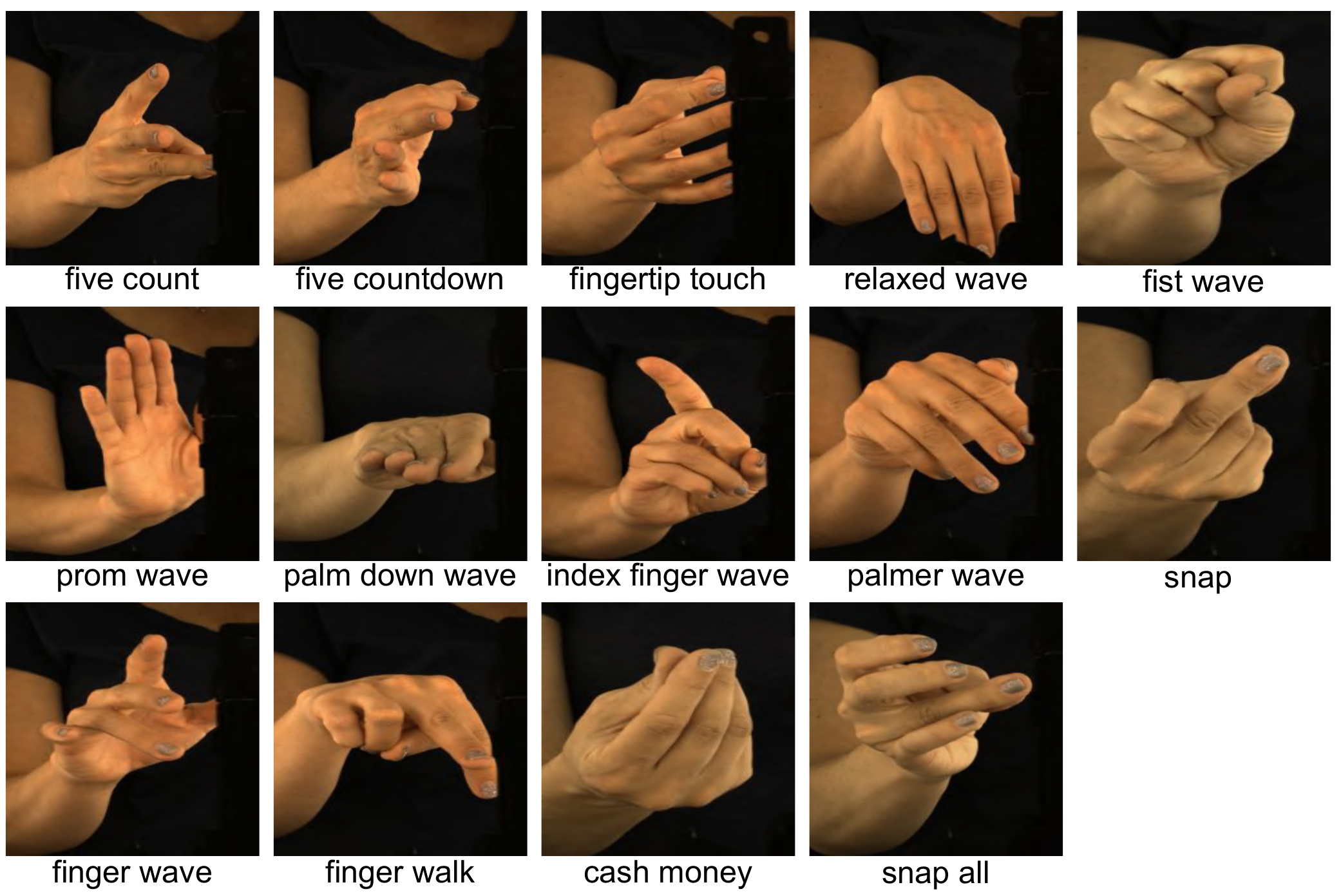}
\end{center}
\vspace*{-5mm}
   \caption{Visualization of the single hand ROM sequences.}
\vspace*{-3mm}
\label{fig:single_rom}
\end{figure}

Figure~\ref{fig:single_rom} shows ROM of single hand sequences. 
Below are detailed descriptions of each sequence.
\begin{itemize}
\item five count: count from one to five. 
\item five countdown: count from five to one.
\item fingertip touch: thumb touch each fingertip.
\item relaxed wave: wrist relaxed, fingertips facing down and relaxed, wave.
\item fist wave: rotate wrist while hand in a fist shape.
\item prom wave: wave with fingers together.
\item palm down wave: wave hand with the palm facing down.
\item index finger wave: hand gesture that represents \enquote{no} sign.
\item palmer wave: palm down, scoop towards you, like petting an animal.
\item snap: snap middle finger and thumb.
\item finger wave: palm down, move fingers like playing the piano.
\item finger walk: mimicking a walking person by index and middle finger.
\item cash money: rub thumb on the index and middle fingertips.
\item snap all: snap each finger on the thumb.
\end{itemize}

\begin{figure}
\begin{center}
\includegraphics[width=1.0\linewidth]{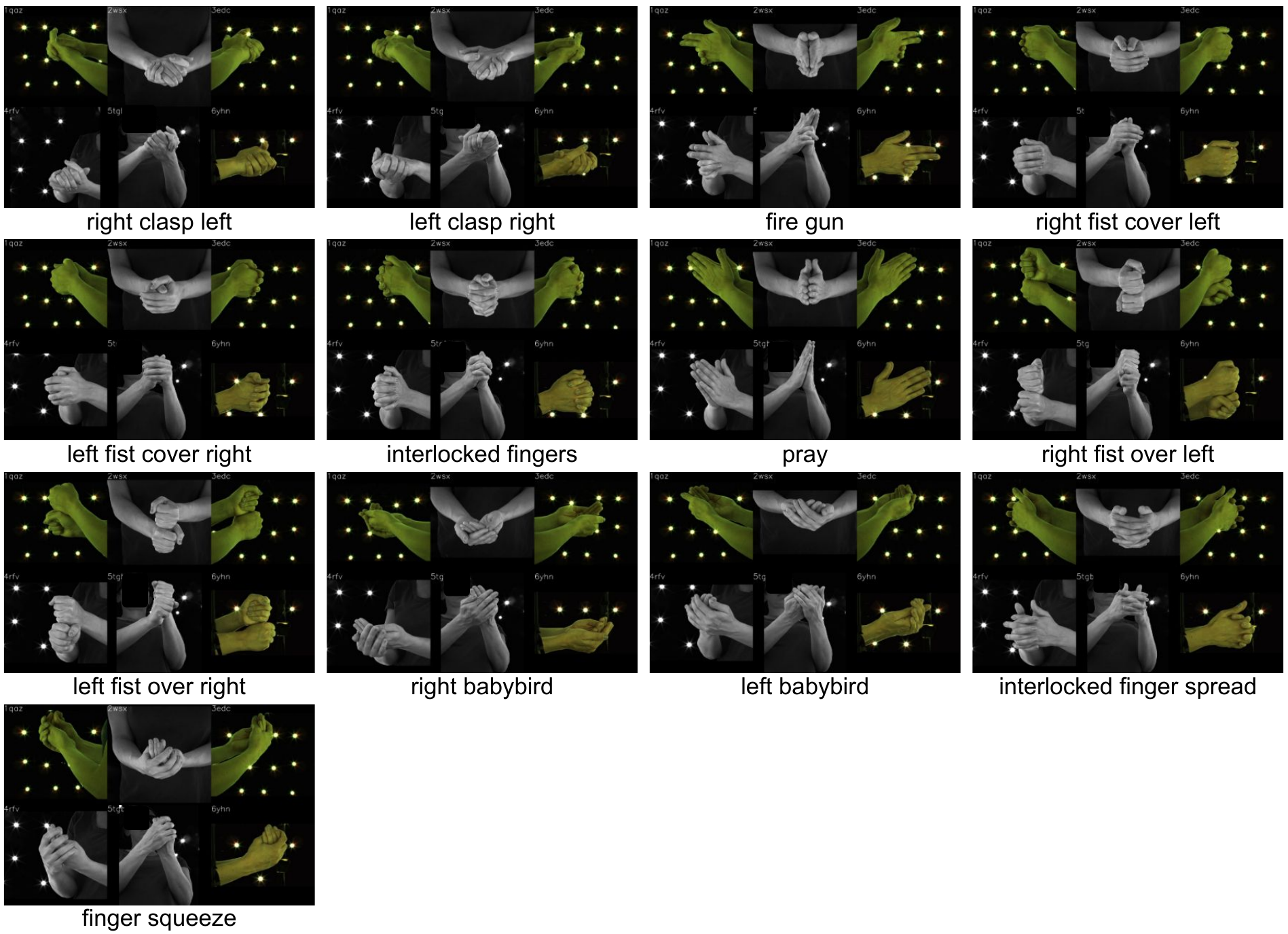}
\end{center}
\vspace*{-5mm}
   \caption{Visualization of the interacting hand PP sequences.}
\vspace*{-3mm}
\label{fig:inter_pp}
\end{figure}

\noindent\textbf{Interacting hand sequences.}
Figure~\ref{fig:inter_pp} shows PP of interacting hand sequences. 
Below are detailed descriptions of each sequence.
\begin{itemize}
\item right clasp left: two hands clasp each other, right hand is on top of the left hand.
\item left clasp right: two hands clasp each other, left hand is on top of the right hand.
\item fire gun: mimicking a gun with two hands together.
\item right fist cover left: right fist completely covers the left hand.
\item left fist cover right: left fist completely covers the right hand.
\item interlocked fingers: fingers of the right and left hands are interlocked.
\item pray: hand sign language that represents praying.
\item right fist over left: right fist is on top of the left fist.
\item left fist over right: left fist is on top of the right fist.
\item right babybird: mimicking caring a babybird with two hands, the right hand is placed at the bottom.
\item left babybird: mimicking caring a babybird with two hands, the left hand is placed at the bottom.
\item interlocked finger spread: fingers of the right and left hands are interlocked yet spread. 
\item finger squeeze: squeeze all five fingers with the other hand.
\end{itemize}

\begin{figure}
\begin{center}
\includegraphics[width=1.0\linewidth]{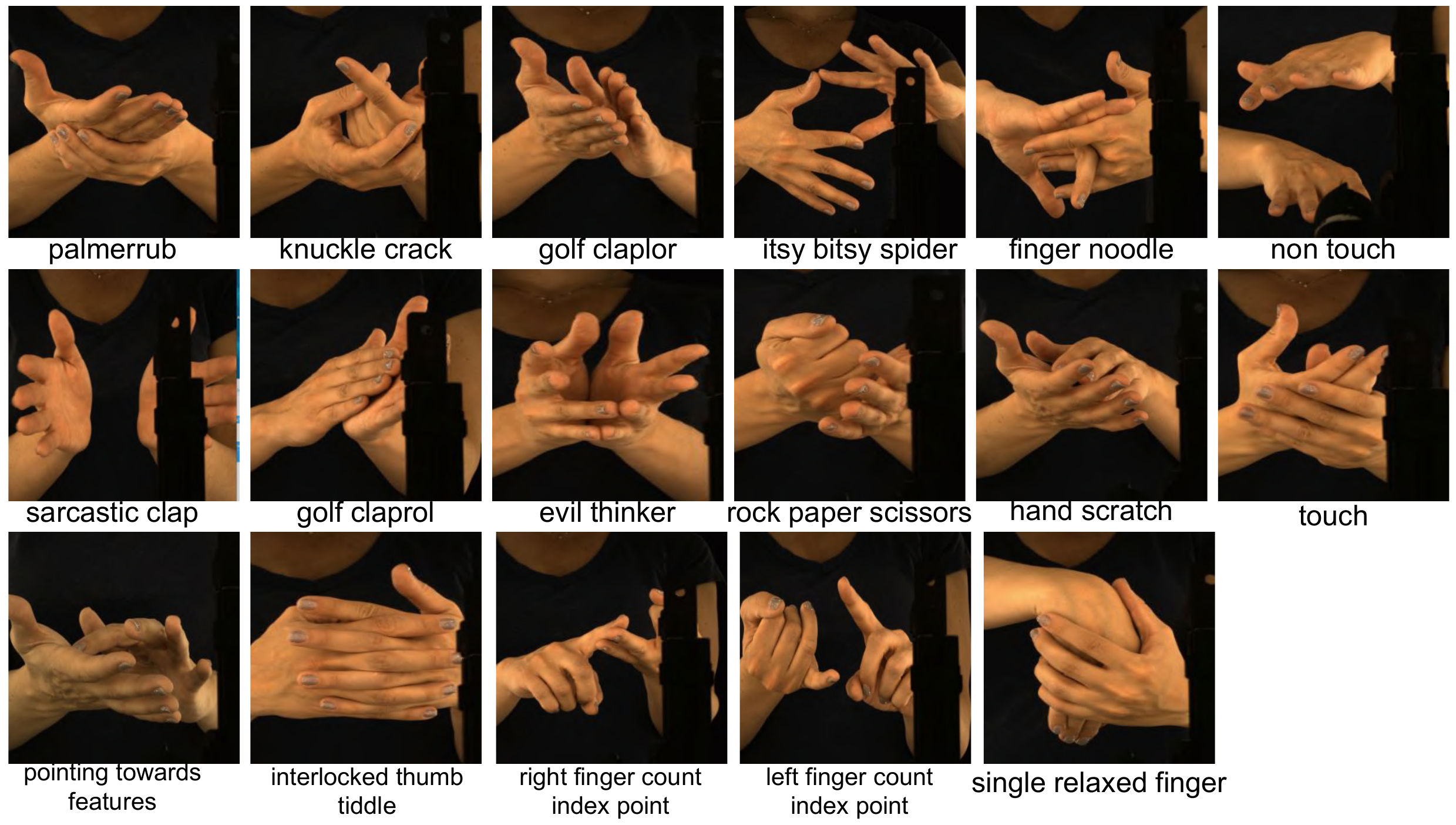}
\end{center}
\vspace*{-5mm}
   \caption{Visualization of the interacting hand ROM sequences.}
\vspace*{-3mm}
\label{fig:ih_rom}
\end{figure}

Finally, Figure~\ref{fig:ih_rom} shows ROM of interacting hand sequences.
Below are detailed descriptions of each sequence.
\begin{itemize}
\item palmerrub: rub palm of one hand with opposite hand's thumb.
\item knuckle crack: crack each finger by having the opposite hand compress a bent finger.
\item golf claplor: light clap, left over right.
\item itsy bitsy spider: finger motion used when singing the children song \enquote{itsy bitsy spider}, like this \href{https://www.youtube.com/watch?v=anjuuQBHzI0}{(link)}.
\item finger noodle: fingers interlocked, palms facing opposite directions, wiggle middle fingers.
\item nontouch: two hands random motion, hands do not touch.
\item sarcastic clap: exaggerated, slow clap.
\item golf claprol: light clap, right over left.
\item evil thinker: wrist together, tap fingers together one at a time.
\item rock paper scissors: hold rock, then paper, then scissors.
\item hand scratch: using the opposite hand, lightly scratch palm then top of hand; switch and do the same with the other hand.
\item touch: two hands interacting randomly in a frame, touching.
\item pointing towards features: using the opposite index finger, point out features on the palm and back of the hand (trace lifelines/wrinkles, etc.).
\item interlocked thumb tiddle: interlock fingers, rotate thumbs around each other.
\item right finger count index point: using the right pointer finger, count up to five on the left hand, starting with the pinky.
\item left finger count index point: using left pointer finger, count up to five on the right hand, starting with the pinky.
\item single relaxed finger: this consists of a series of actions: (1) touch each fingertip to the center of the palm for the same hand, do this for both hands, (2)
interlock fingers and press palms out, (3) with the opposite hand, hold wrist, 
(4) with the opposite hand, bend wrist down and back, 
(5) point at watch on both wrists,
(6) circle wrists,
(7) look at nails,
and (8) point at yourself with thumbs then with index fingers.
\end{itemize}

\newpage

\section{InterHand2.6M human annotation procedure}
\begin{figure}
\begin{center}
\includegraphics[width=0.9\linewidth]{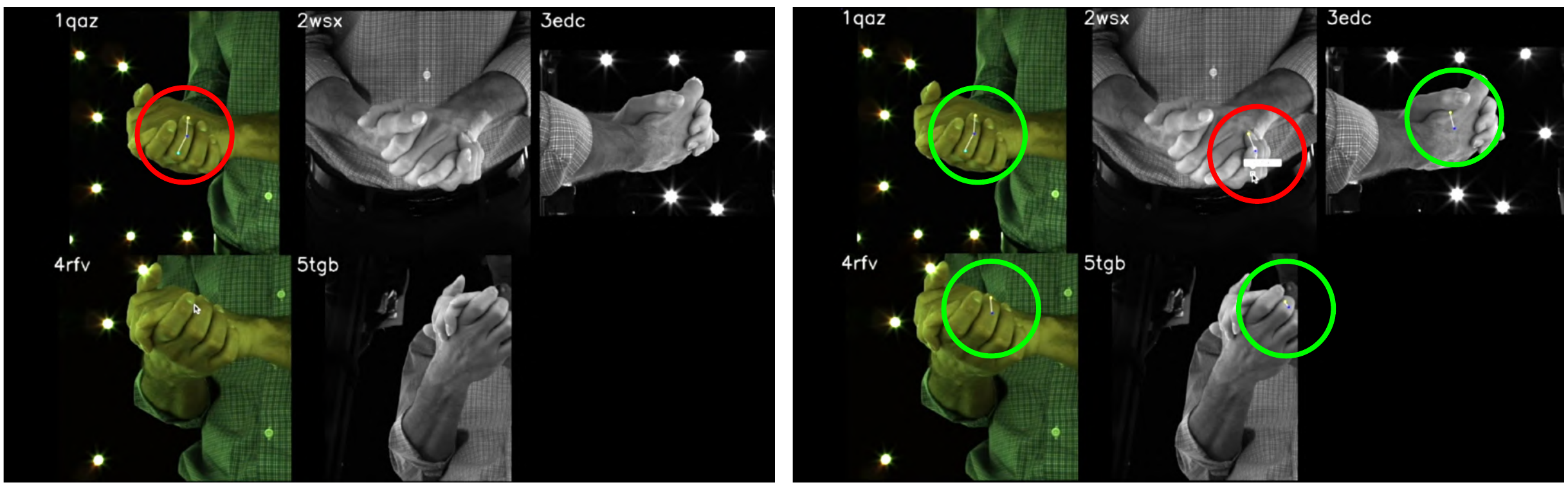}
\end{center}
\vspace*{-5mm}
   \caption{The human annotation procedure of InterHand2.6M.}
\vspace*{-5mm}
\label{fig:human_annotation}
\end{figure}

Figure~\ref{fig:human_annotation} shows human annotation procedure of InterHand2.6M.
In the left figure, an annotator clicks hand joint positions at the easiest view (red circle).
Then, the annotator clicks the positions of the same hand joints at another view (red circle).
Our human annotation tool automatically triangulates human annotations from two views in the 3D space and projects the 3D point to remaining views (green circles).

\section{InterHand2.6M capture studio environment}
\begin{figure}
\begin{center}
\includegraphics[width=0.8\linewidth]{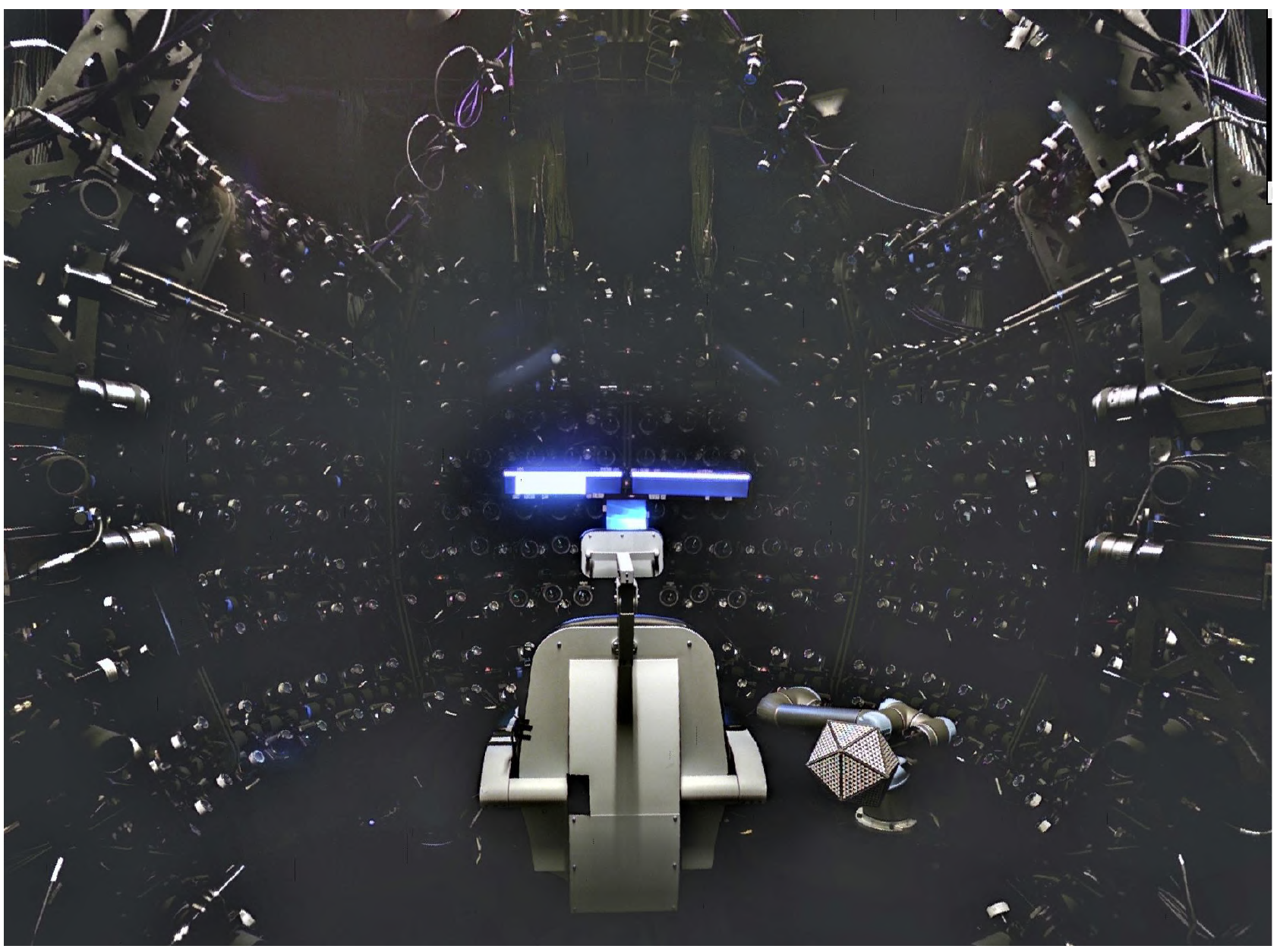}
\end{center}
\vspace*{-5mm}
   \caption{Rendering of our constructed multi-view studio.}
\vspace*{-3mm}
\label{fig:studio}
\end{figure}

Figure~\ref{fig:studio} shows a rendering of our constructed multi-view studio for the data capture.

\clearpage
 
\section{Evaluation on various test set using various training set}
We provide more experimental results on various test set of the InterHand2.6M (\textit{i.e.}, Val (M), Test (H), Test (M), and Test (H+M)) by training InterNet on various training set (\textit{i.e.}, Train (H), Train (M), and Train (H+M)). 
Table~\ref{table:ap_h_comparison} and Table~\ref{table:mrrpe_comparison} show $\mathrm{AP_{h}}$ and MRRPE on all various testing sets from models trained on different set.
Table~\ref{table:mpjpe_val}, ~\ref{table:mpjpe_test_human}, ~\ref{table:mpjpe_test_machine}, and ~\ref{table:mpjpe_test_all} show MPJPE on Val (M), Test (H), Test (M), and Test (H+M), respectively.
T, I, M, R, and P denote thumb, index, middle, ring, and pinky fingers, respectively. 
For each finger, 1 is finger root (closest to the wrist) and the next indices toward fingertip. 
We averaged errors of right and left hand.

\begin{table}
\centering
\setlength\tabcolsep{1.0pt}
\def\arraystretch{1.1}
\begin{tabular}{C{2.5cm}C{2.0cm}C{2.0cm}C{2.0cm}C{2.0cm}}
\specialrule{.1em}{.05em}{.05em}
training set & Val (M) & Test (H) & Test (M) & Test (H+M) \\ \hline
Train (H) & \textbf{99.10} & \textbf{99.79} & 98.95 & 99.01 \\
Train (M) & 97.79 & 99.07 & 98.85 & 98.87 \\
Train (H+M) & 98.14 & 99.77 & \textbf{99.03} & \textbf{99.09}\\ \hline
\end{tabular}
\caption{$\mathrm{AP_{h}}$ comparison from models trained with different training set.}
\label{table:ap_h_comparison}
\end{table}

\begin{table}
\centering
\setlength\tabcolsep{1.0pt}
\def\arraystretch{1.1}
\begin{tabular}{C{2.5cm}C{2.0cm}C{2.0cm}C{2.0cm}C{2.0cm}}
\specialrule{.1em}{.05em}{.05em}
training set & Val (M) & Test (H) & Test (M) & Test (H+M) \\ \hline
Train (H) & 40.06 & 21.80 & 38.50 & 36.80 \\
Train (M) & 40.50 & 23.21 & 38.84 & 37.25 \\
Train (H+M) & \textbf{35.72} & \textbf{20.26} & \textbf{33.97} & \textbf{32.57}\\ \hline
\end{tabular}
\caption{MRRPE comparison from models trained with different training set.}
\label{table:mrrpe_comparison}
\end{table}

\begin{table}
\centering
\setlength\tabcolsep{1.0pt}
\def\arraystretch{1.1}
\scalebox{0.65}{
\begin{tabular}{L{2.3cm}C{0.7cm}C{0.7cm}C{0.7cm}C{0.7cm}C{0.7cm}C{0.7cm}C{0.7cm}C{0.7cm}C{0.7cm}C{0.7cm}C{0.7cm}C{0.7cm}C{0.7cm}C{0.7cm}C{0.7cm}C{0.7cm}C{0.7cm}C{0.7cm}C{0.7cm}C{0.7cm}C{0.9cm}}
\specialrule{.1em}{.05em}{.05em}
training set & T4 & T3 & T2 & T1 & I4 & I3 & I2 & I1 & M4 & M3 & M2 & M1 & R4 & R3 & R2 & R1 & P4 & P3 & P2 & P1 & avg. \\ \hline
\multicolumn{22}{l}{\textbf{\textit{results on single hand sequences}}}  \\
Train (H) & 18.5 & 14.5 & 11.1 & 7.7 & 20.6 & 17.7 & 15.3 & 12.0 & 22.9 & 19.7 & 16.7 & 12.1 & 21.5 & 18.0 & \textbf{15.0} & 11.2 & \textbf{19.6} & \textbf{16.7} & \textbf{14.4} & \textbf{10.3} & 15.02 \\            
Train (M) & 18.2 & 14.5 & 11.1 & \textbf{7.2} & 20.8 & 18.0 & 15.9 & 12.3 & 23.0 & 20.4 & 17.5 & 12.3 & 21.7 & 18.8 & 15.7 & 11.4 & 20.5 & 17.7 & 15.1 & 10.5 & 15.36 \\    
Train (H+M) & \textbf{17.6} & \textbf{14.0} & \textbf{10.7} & \textbf{7.2} & \textbf{19.7} & \textbf{17.2} & \textbf{15.1} & \textbf{11.7} & \textbf{21.5} & \textbf{19.0} & \textbf{16.4} & \textbf{11.8} & \textbf{20.4} & \textbf{17.8} & \textbf{15.0} & \textbf{10.9} & \textbf{19.6} & 17.1 & 14.7 & \textbf{10.3} & \textbf{14.65} \\ \hline
\multicolumn{22}{l}{\textbf{\textit{results on interacting hand sequences}}}  \\
Train (H) &  25.9 & 18.8 & 15.3 & 10.6 & 28.8 & 24.3 & 20.7 & 15.8 & 29.7 & 24.2 & 21.3 & 15.8 & 26.5 & 22.0 & 19.3 & 14.9 & 25.2 & 21.2 & 18.8 & 14.4 & 19.70 \\            
Train (M) &  25.9 & 18.8 & 15.4 & 10.1 & 32.2 & 26.1 & 21.6 & 15.8 & 29.7 & 24.7 & 21.6 & 15.9 & 27.0 & 22.3 & 19.5 & 14.9 & 26.0 & 21.7 & 19.1 & 14.3 & 20.13 \\            
Train (H+M) & \textbf{23.8} & \textbf{17.5} & \textbf{14.2} & \textbf{9.7} & \textbf{28.0} & \textbf{23.4} & \textbf{19.6} & \textbf{14.5} & \textbf{27.8} & \textbf{22.9} & \textbf{20.1} & \textbf{14.6} & \textbf{24.9} & \textbf{20.8} & \textbf{18.2} & \textbf{13.9} & \textbf{24.2} & \textbf{20.4} & \textbf{18.0} & \textbf{13.5} & \textbf{18.58} \\ \hline
\end{tabular}
}
\caption{MPJPE of our InterNet on the Val (M) of InterHand2.6M dataset using various training set.}
\vspace*{-3mm}
\label{table:mpjpe_val}
\end{table}

\begin{table}
\centering
\setlength\tabcolsep{1.0pt}
\def\arraystretch{1.1}
\scalebox{0.65}{
\begin{tabular}{L{2.3cm}C{0.7cm}C{0.7cm}C{0.7cm}C{0.7cm}C{0.7cm}C{0.7cm}C{0.7cm}C{0.7cm}C{0.7cm}C{0.7cm}C{0.7cm}C{0.7cm}C{0.7cm}C{0.7cm}C{0.7cm}C{0.7cm}C{0.7cm}C{0.7cm}C{0.7cm}C{0.7cm}C{0.9cm}}
\specialrule{.1em}{.05em}{.05em}
training set & T4 & T3 & T2 & T1 & I4 & I3 & I2 & I1 & M4 & M3 & M2 & M1 & R4 & R3 & R2 & R1 & P4 & P3 & P2 & P1 & avg. \\ \hline
\multicolumn{22}{l}{\textbf{\textit{results on single hand sequences}}}  \\
Train (H) & 13.4 & 11.1 & 9.3 & 7.7 & 13.1 & 11.8 & 10.8 & 9.1 & 13.9 & 11.7 & 10.7 & 9.1 & 14.0 & 11.2 & 10.2 & 9.0 & 13.2 & 11.1 & 10.0 & 8.5 & 10.42 \\
Train (M) & 13.1 & 11.1 & 9.2 & 7.4 & 13.5 & 12.3 & 11.3 & 9.3 & 14.2 & 12.4 & 11.2 & 9.4 & 13.9 & 11.5 & 10.6 & 9.4 & 13.1 & 11.2 & 10.4 & 8.8 & 10.64 \\
Train (H+M) & \textbf{12.1} & \textbf{10.4} & \textbf{8.9} & \textbf{7.3} & \textbf{12.1} & \textbf{11.2} & \textbf{10.2} & \textbf{8.6} & \textbf{13.0} & \textbf{11.2} & \textbf{10.2} & \textbf{8.7} & \textbf{12.8} & \textbf{10.6} & \textbf{9.7} & \textbf{8.7} & \textbf{12.4} & \textbf{10.6} & \textbf{9.8} & \textbf{8.2} & \textbf{9.85} \\ \hline
\multicolumn{22}{l}{\textbf{\textit{results on interacting hand sequences}}}  \\
Train (H) & 18.4 & 14.5 & 11.7 & 10.0 & 17.2 & 14.7 & 13.3 & 11.7 & 17.1 & 14.6 & 13.0 & 11.3 & 16.4 & 14.1 & 12.4 & 10.7 & 15.7 & 13.7 & 12.5 & 11.0 & 13.05 \\
Train (M) & 20.2 & 15.7 & 12.3 & 10.2 & 19.7 & 16.6 & 14.7 & 12.4 & 19.4 & 16.3 & 14.2 & 12.0 & 18.3 & 15.4 & 13.5 & 11.5 & 17.0 & 14.9 & 13.4 & 11.7 & 14.26 \\
Train (H+M) & \textbf{17.1} & \textbf{13.6} & \textbf{11.0} & \textbf{9.6} & \textbf{16.1} & \textbf{13.8} & \textbf{12.5} & \textbf{11.0} & \textbf{16.0} & \textbf{13.8} & \textbf{12.2} & \textbf{10.7} & \textbf{15.3} & \textbf{13.2} & \textbf{11.7} & \textbf{10.2} & \textbf{14.9} & \textbf{13.1} & \textbf{11.8} & \textbf{10.4} & \textbf{12.29} \\ \hline
\end{tabular}
}
\caption{MPJPE of our InterNet on the Test (H) of InterHand2.6M dataset using various training set.}
\vspace*{-3mm}
\label{table:mpjpe_test_human}
\end{table}

\begin{table}
\centering
\setlength\tabcolsep{1.0pt}
\def\arraystretch{1.1}
\scalebox{0.65}{
\begin{tabular}{L{2.3cm}C{0.7cm}C{0.7cm}C{0.7cm}C{0.7cm}C{0.7cm}C{0.7cm}C{0.7cm}C{0.7cm}C{0.7cm}C{0.7cm}C{0.7cm}C{0.7cm}C{0.7cm}C{0.7cm}C{0.7cm}C{0.7cm}C{0.7cm}C{0.7cm}C{0.7cm}C{0.7cm}C{0.9cm}}
\specialrule{.1em}{.05em}{.05em}
training set & T4 & T3 & T2 & T1 & I4 & I3 & I2 & I1 & M4 & M3 & M2 & M1 & R4 & R3 & R2 & R1 & P4 & P3 & P2 & P1 & avg. \\ \hline
\multicolumn{22}{l}{\textbf{\textit{results on single hand sequences}}}  \\
Train (H) & 16.6 & 12.8 & 9.7 & 6.9 & 17.2 & 14.7 & 13.0 & 11.0 & 18.6 & 15.8 & 13.7 & 10.6 & 17.7 & 14.7 & \textbf{12.9} & 10.0 & 16.2 & \textbf{13.6} & \textbf{12.0} & 9.7 & 12.74 \\
Train (M) & \textbf{15.7} & \textbf{12.1} & \textbf{9.3} & \textbf{6.5} & 16.6 & 14.5 & 13.1 & 10.8 & 18.0 & 15.7 & 13.8 & 10.4 & 17.4 & 14.8 & 13.1 & 10.1 & 16.2 & 13.8 & 12.2 & 9.6 & 12.56 \\
Train  (H+M) & \textbf{15.7} & \textbf{12.1} & \textbf{9.3} & 6.6 & \textbf{16.1} & \textbf{14.2} & \textbf{12.8} & \textbf{10.6} & \textbf{17.4} & \textbf{15.3} & \textbf{13.5} & \textbf{10.3} & \textbf{16.9} & \textbf{14.5} & \textbf{12.9} & \textbf{9.8} & \textbf{15.8} & \textbf{13.6} & \textbf{12.0} & \textbf{9.4} & \textbf{12.32} \\ \hline
\multicolumn{22}{l}{\textbf{\textit{results on interacting hand sequences}}}  \\
Train (H) & 23.3 & 17.8 & 13.8 & 9.5 & 25.9 & 21.8 & 18.8 & 15.0 & 29.5 & 22.3 & 18.7 & 14.2 & 25.8 & 19.8 & 17.5 & 13.7 & 23.5 & 18.9 & 16.7 & 13.6 & 18.10 \\
Train (M) & 23.9 & 18.0 & 13.8 & 9.0 & 28.8 & 23.3 & 19.5 & 15.2 & 29.4 & 22.9 & 19.1 & 14.5 & 25.6 & 20.4 & 17.9 & 14.0 & 24.3 & 19.6 & 17.3 & 13.9 & 18.59 \\
Train (H+M) & \textbf{21.5} & \textbf{16.4} & \textbf{12.6} & \textbf{8.5} & \textbf{24.4} & \textbf{20.5} & \textbf{17.4} & \textbf{13.7} & \textbf{27.7} & \textbf{20.9} & \textbf{17.5} & \textbf{13.1} & \textbf{24.0} & \textbf{18.6} & \textbf{16.4} & \textbf{12.7} & \textbf{22.3} & \textbf{17.9} & \textbf{15.8} & \textbf{12.6} & \textbf{16.88} \\ \hline
\end{tabular}
}
\caption{MPJPE of our InterNet on the Test (M) of InterHand2.6M dataset using various training set.}
\vspace*{-3mm}
\label{table:mpjpe_test_machine}
\end{table}

\begin{table}
\centering
\setlength\tabcolsep{1.0pt}
\def\arraystretch{1.1}
\scalebox{0.65}{
\begin{tabular}{L{2.3cm}C{0.7cm}C{0.7cm}C{0.7cm}C{0.7cm}C{0.7cm}C{0.7cm}C{0.7cm}C{0.7cm}C{0.7cm}C{0.7cm}C{0.7cm}C{0.7cm}C{0.7cm}C{0.7cm}C{0.7cm}C{0.7cm}C{0.7cm}C{0.7cm}C{0.7cm}C{0.7cm}C{0.9cm}}
\specialrule{.1em}{.05em}{.05em}
training set & T4 & T3 & T2 & T1 & I4 & I3 & I2 & I1 & M4 & M3 & M2 & M1 & R4 & R3 & R2 & R1 & P4 & P3 & P2 & P1 & avg. \\ \hline
\multicolumn{22}{l}{\textbf{\textit{results on single hand sequences}}}  \\
Train (H) & 16.4 & 12.7 & 9.7 & 7.0 & 16.9 & 14.5 & 12.9 & 10.8 & 18.3 & 15.5 & 13.5 & 10.5 & 17.5 & 14.5 & \textbf{12.7} & 9.9 & 16.0 & \textbf{13.4} & \textbf{11.9} & 9.6 & 12.58 \\
Train (M) & 15.5 & \textbf{12.0} & \textbf{9.3} & \textbf{6.5} & 16.4 & 14.4 & 13.0 & 10.7 & 17.7 & 15.5 & 13.7 & 10.3 & 17.2 & 14.6 & 12.9 & 10.1 & 16.0 & 13.7 & 12.1 & 9.6 & 12.43 \\
Train (H+M) & \textbf{15.4} & \textbf{12.0} & \textbf{9.3} & 6.7 & \textbf{15.8} & \textbf{14.0} & \textbf{12.6} & \textbf{10.4} & \textbf{17.1} & \textbf{15.0} & \textbf{13.3} & \textbf{10.2} & \textbf{16.6} & \textbf{14.3} & \textbf{12.7} & \textbf{9.7} & \textbf{15.6} & \textbf{13.4} & \textbf{11.9} & \textbf{9.4} & \textbf{12.16} \\ \hline
\multicolumn{22}{l}{\textbf{\textit{results on interacting hand sequences}}}  \\
Train (H) & 22.4 & 17.1 & 13.4 & 9.7 & 24.4 & 20.5 & 17.8 & 14.4 & 27.1 & 20.9 & 17.7 & 13.7 & 23.8 & 18.8 & 16.6 & 13.1 & 22.0 & 18.0 & 15.9 & 13.1 & 17.16 \\
Train (M) & 23.2 & 17.6 & 13.6 & 9.3 & 27.2 & 22.1 & 18.6 & 14.7 & 27.5 & 21.7 & 18.2 & 14.0 & 24.0 & 19.5 & 17.1 & 13.5 & 22.9 & 18.8 & 16.5 & 13.5 & 17.79 \\
Train (H+M) & \textbf{20.7} & \textbf{15.9} & \textbf{12.3} & \textbf{8.8} & \textbf{23.0} & \textbf{19.3} & \textbf{16.6} & \textbf{13.2} & \textbf{25.4} & \textbf{19.5} & \textbf{16.5} & \textbf{12.6} & \textbf{22.1} & \textbf{17.6} & \textbf{15.5} & \textbf{12.2} & \textbf{20.9} & \textbf{17.1} & \textbf{15.0} & \textbf{12.2} & \textbf{16.02} \\ \hline
\end{tabular}
}
\caption{MPJPE of our InterNet on the Test (H+M) of InterHand2.6M dataset using various training set.}
\vspace*{-3mm}
\label{table:mpjpe_test_all}
\end{table}

\clearpage

\section{Qualitative results}
\begin{figure}
\begin{center}
\includegraphics[width=1.0\linewidth]{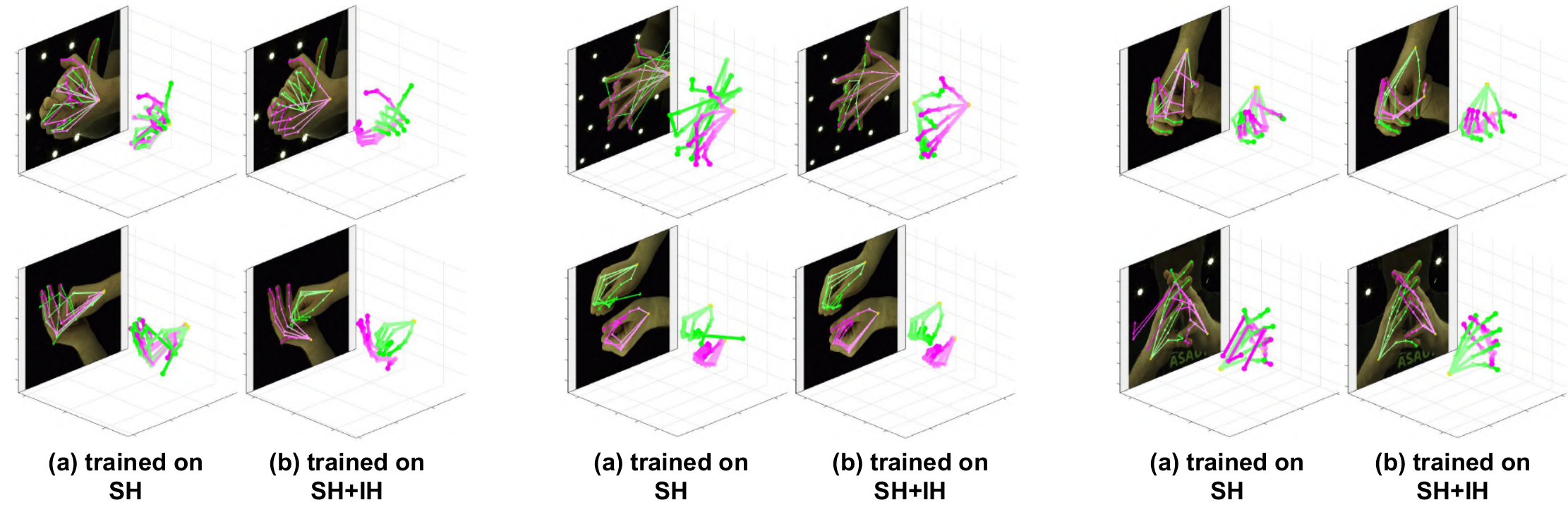}
\end{center}
\vspace*{-7mm}
   \caption{Qualitative results comparison of our InterNet trained on (a) only single hand data and (b) both single and interacting hand data from the proposed InterHand2.6M dataset.}
\vspace*{-6mm}
\label{fig:sh_ih_qualitative_compare}
\end{figure}

\begin{figure}
\begin{center}
\includegraphics[width=1.0\linewidth]{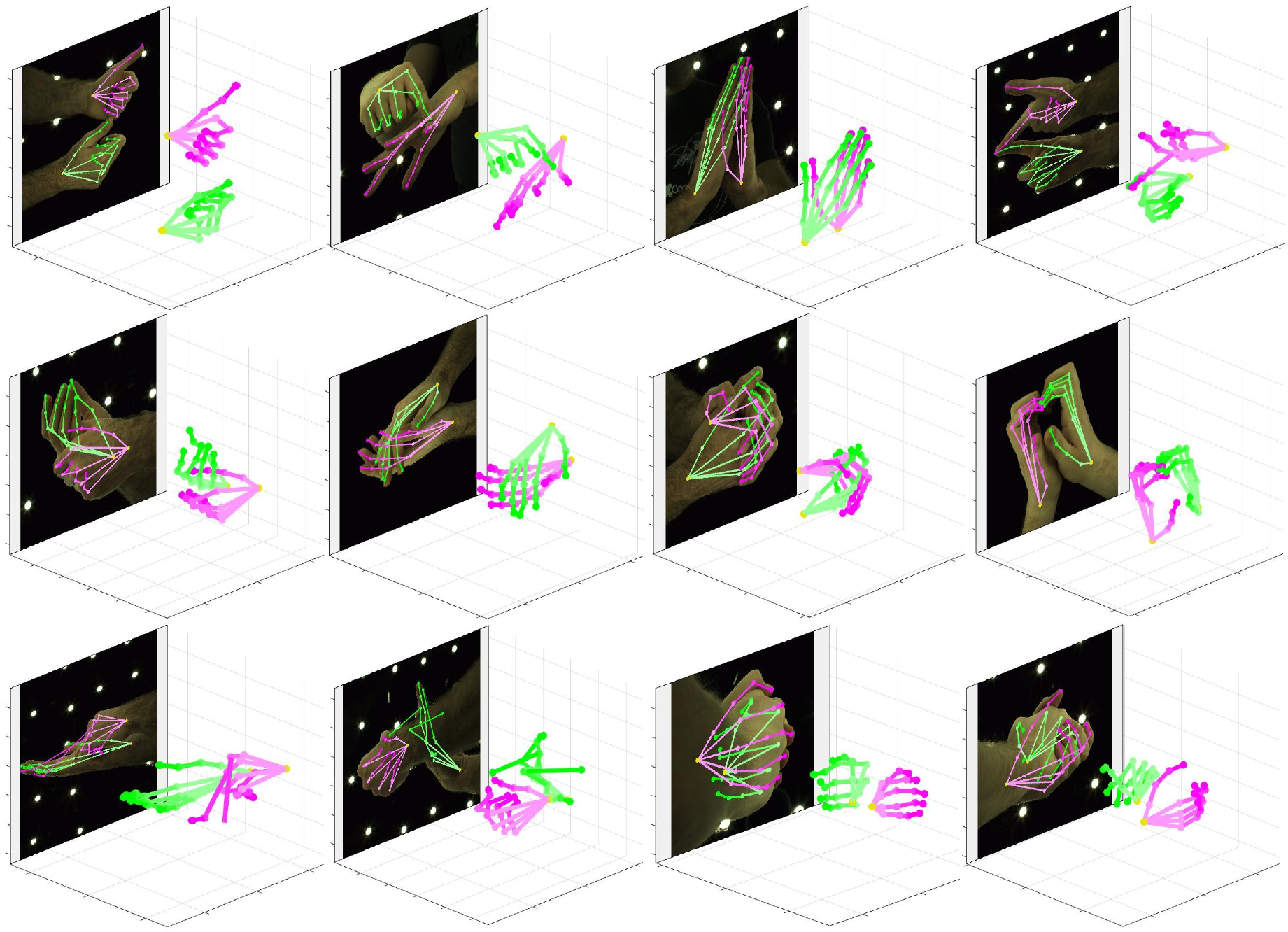}
\end{center}
\vspace*{-7mm}
   \caption{Qualitative results (top two rows) and failure case (last row) of our InterNet on the proposed InterHand2.6M dataset.}
\vspace*{-3mm}
\label{fig:qualitative_failure}
\end{figure}

We compare the qualitative results of our InterNet trained on (a) only single hand data and (b) both single and interacting hand data in Figure~\ref{fig:sh_ih_qualitative_compare}. 
As the figure shows, when InterNet is trained only on single hand data, it provides a reasonable 3D hand pose when the input image contained separated two hands (\textit{i.e.}, bottom middle example).
However, it totally fails for all interacting hand sequences.
We provide more qualitative results and failure cases of our InterNet on Test (H+M) of the proposed InterHand2.6M dataset in Figure~\ref{fig:qualitative_failure}.
As the figure shows, severe occlusions make 2.5D hand pose estimation and right hand-relative left hand depth estimation fail.

\clearpage